\documentclass[lettersize,journal]{IEEEtran}
\usepackage{amsmath,amsfonts,amssymb}
\usepackage{algorithmic}
\usepackage{algorithm}
\usepackage{array}
\usepackage[caption=false,font=normalsize,labelfont=sf,textfont=sf]{subfig}
\usepackage{textcomp}
\usepackage{stfloats}
\usepackage{url}
\usepackage{verbatim}
\usepackage{graphicx}
\usepackage{cite}
\usepackage{color}
\usepackage{multirow}
\usepackage{amsthm}
\usepackage{threeparttable}
\newcommand{\ie}{i.e.\ }
\newcommand{\etal}{\emph{et al.}\ }
\begin{document}

\title{MS-DETR: Multispectral Pedestrian Detection Transformer with Loosely Coupled Fusion and Modality-Balanced Optimization}

\author{Yinghui Xing,~\IEEEmembership{Member,~IEEE}, Shuo Yang, Song Wang, Shizhou Zhang, Guoqiang Liang, Xiuwei Zhang, and Yanning Zhang,~\IEEEmembership{Senior Member,~IEEE}

\thanks{This work was supported in part by the National Natural Science Foundation of China (NFSC) under Grant 62201467 and Grant 62101453; in part by the Guangdong Basic and Applied Basic Research Foundation under Grant 2021A1515110544; in part by the Project funded by China Postdoctoral Science Foundation under Grant 2022TQ0260 and Grant 2023M742842; in part by the Young Talent Fund of Xi'an Association for Science and Technology under Grant 959202313088; in part by the Innovation Capability Support Program of Shaanxi (No. 2024ZC-KJXX-043), and in part by the Natural Science Basic Research Program of Shaanxi Provience (No. 2022JC-DW-08, and 2024JC-YBQN-0719).~\textit{(Corresponding author: Shizhou Zhang.)}}
\thanks{Yinghui Xing, Shuo Yang, Song Wang, Shizhou Zhang, Guoqiang Liang, Xiuwei Zhang and Yanning Zhang are with the School of Computer Science, Northwestern Polytechnical University, Xi’an 710072, China. Yinghui Xing is also with the Research Development Institute of Northwestern Polytechnical University in
Shenzhen, Shenzhen 518057, China. (e-mail: xyh\_7491@nwpu.edu.cn).}
}


\maketitle

\begin{abstract}
Multispectral pedestrian detection is an important task for many around-the-clock applications, since the visible and thermal modalities can provide complementary information especially under low light conditions. 
Due to the presence of two modalities, misalignment and modality imbalance are the most significant issues in multispectral pedestrian detection.  
In this paper, we propose \textbf{M}ulti\textbf{S}pectral pedestrian \textbf{DE}tection \textbf{TR}ansformer (MS-DETR) to fix above issues. 
MS-DETR consists of two modality-specific backbones and Transformer encoders, followed by a multi-modal Transformer decoder, and the visible and thermal features are fused in the multi-modal Transformer decoder. 
To well resist the misalignment between multi-modal images, we design a loosely coupled fusion strategy by sparsely sampling some keypoints from multi-modal features independently and fusing them with adaptively learned attention weights.
Moreover, based on the insight that not only different modalities, but also different pedestrian instances tend to have different confidence scores to final detection, we further propose an instance-aware modality-balanced optimization strategy, which preserves visible and thermal decoder branches and aligns their predicted slots through an instance-wise dynamic loss. 
Our end-to-end MS-DETR shows superior performance on the challenging KAIST, CVC-14 and LLVIP benchmark datasets.
The source code is available at https://github.com/YinghuiXing/MS-DETR.

\end{abstract}


\section{Introduction}
\IEEEPARstart{P}{edestrian} detection is a popular research topic in autonomous driving~\cite{enzweiler2008monocular}, automated video surveillance~\cite{wang2013scene}, and robotics. 
With the development of deep convolutional neural networks (CNN), many works are proposed~\cite{zhang2016faster, mao2017can, liu2019high, abdelmutalab2022pedestrian, islam2022pedestrian, hsu2023pedestrian}, which greatly boost the development of pedestrian detection. 
However, as visible cameras are sensitive to lighting conditions, even an excellent detector can not accurately detect the pedestrians in adverse environmental conditions~\cite{segments}, limiting many around-the-clock applications. 
To solve this problem, multispectral systems, which have two types of camera sensors, \ie visible and thermal, have been introduced. 
Thermal infrared cameras capture infrared radiation emitted by objects.
They are insensitive to illumination and weather changes, thus can provide complementary information under adverse illumination conditions. 

\begin{figure}[t]
  \centering
  \includegraphics[scale=0.15]{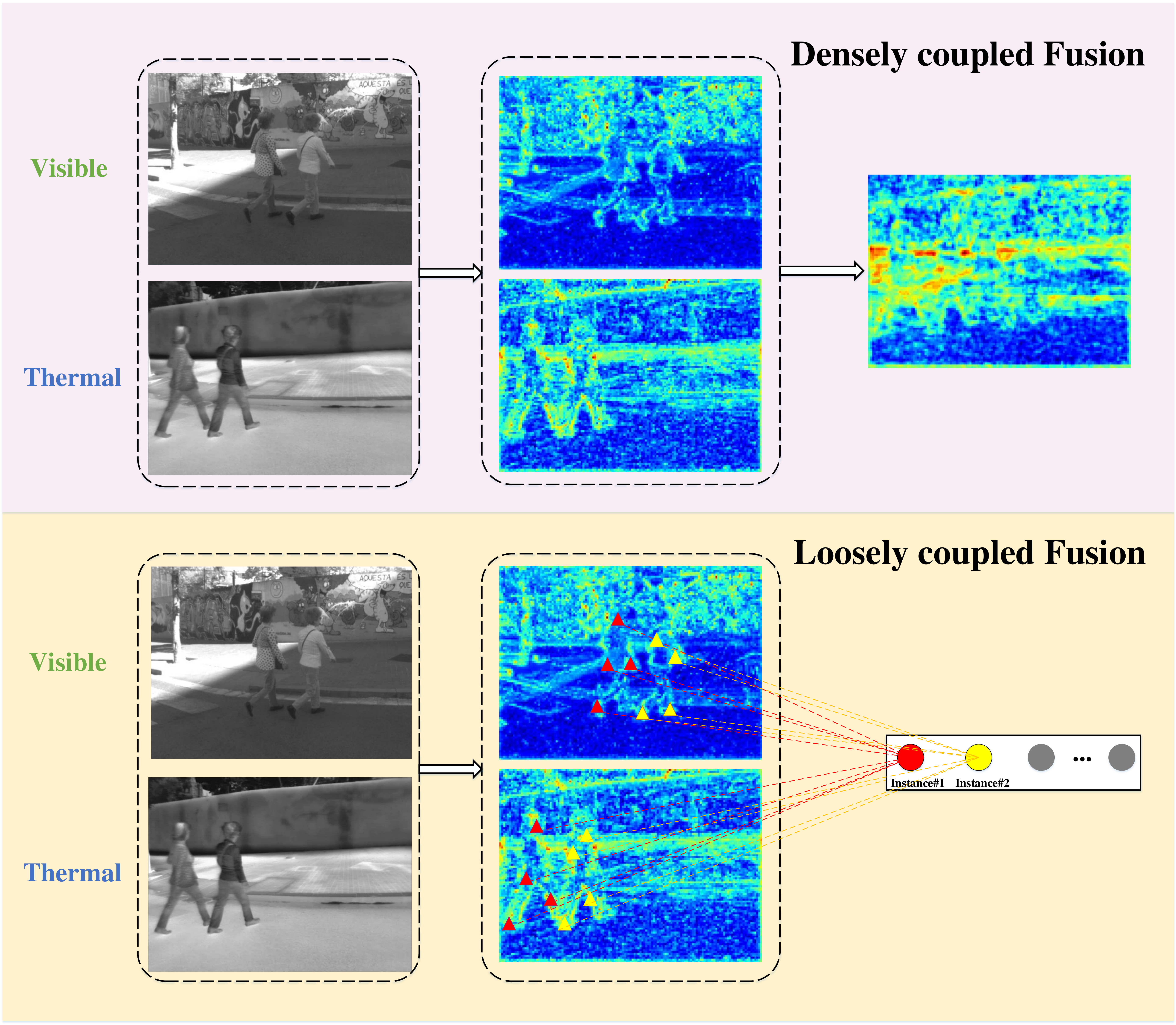}
  \caption{Different fusion strategy of multispectral pedestrian detection. Densely coupled fusion, e.g., concatenation and addition, tends to drift from the target on the misaligned image pairs. In contrast, our loosely coupled fusion aggregates sampled keypoints, which is robust to resist the misalignment.}
  \label{fig:motivation}
\end{figure}

Benefitting from the availablity of KAIST dataset~\cite{KAIST}, some detection methods designed for visible modality are extended to multispectral cases~\cite{KAIST,choi2016multi,konig2017fully,liu2016multispectral,park2018unified,li2019illumination,guan2019fusion, zhang2021guided, dasgupta2022spatio, zhu2023multi}. 
With the development of CNN on object detection, considerable progress has also been made on multispectral pedestrian detection.
However, most of them require some cumbersome modules, like pre-defined anchor boxes~\cite{ren2015faster}, a large scale of proposals~\cite{girshick2014rich} 
and non-maximum suppression (NMS).

\begin{figure*}[t]
  \centering
  \includegraphics[scale=0.13]{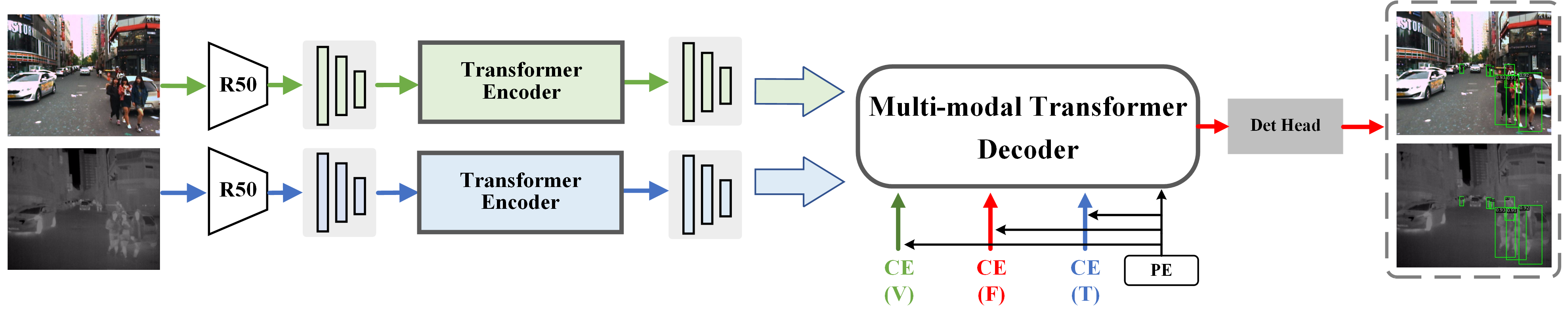}
  \caption{Overall architecture of our MS-DETR, where two modality-specific CNN backbones and two modality-specific Transformer encoders are used for visible and thermal images feature extraction. The multi-modal Transformer decoder takes feature maps, positional encodings (PE) and modality-specific content embeddings (CE) as inputs to generate three sets of prediction slots. V, F, T are acronyms for visible, fusion, and thermal.}
  \label{fig:overall_architecture}
\end{figure*}

Very recently, an end-to-end DEtection TRansformer (DETR)~\cite{DETR} is proposed to reformulate object detection task as a direct set prediction problem. In this paper, we propose MultiSpectral pedestrian DEtection TRansformer (MS-DETR) to extend the end-to-end DETR framework into the field of multispectral pedestrian detection.
Our MS-DETR mainly contains two modality-specific CNN backbones and Transformer encoders, followed with a multi-modal Transformer decoder. 
Firstly, a ResNet-50 network combined with a Transformer encoder is adopted as the modality-specific feature extractor for the visible and thermal modality respectively.
As for the multi-modal decoder, 
conventional densely coupled fusion strategies such as concatenation and addition would require strictly aligned features between the modalities when embedded into the Transformer layers. However, most of multispectral
pedestrian detection datasets, like KAIST and CVC-14, have some misaligned pedestrian pairs between visible and thermal modalities. 
Using such densely coupled fusion strategies may not well aggregate the prominent features, leading to inferior performance, as shown in Fig.~\ref{fig:motivation}.
Therefore, to well suit the multi-modal decoder of MS-DETR, we introduce a loosely coupled fusion strategy instead. By sparsely sampling some keypoints and their corresponding attention weights from the two modalities independently, the multi-modal features are aggregated adaptively, which naturally avoids the strict aligned requirement between visible and thermal modalities. 

From the other hand, during the optimization process of MS-DETR, it may be prone to be dominated by one of the modalities and the detector might suffer from the modality-imbalance problem~\cite{peng2022balanced,wu2022characterizing}. 
Generally speaking, in daytime, pedestrians in visible images include clearer texture features and in nighttime thermal images contain more distinct shape informations. 
While in many cases, pedestrians in thermal images still provide more obvious features than visible images in daytime such as the pedestrians stand in the shadow region or get over-exposed and vice versa.
In other words, not only different modalities tend to obtain different confidence scores, 
but also each different pedestrian instance can have uneven contributions to the object losses under diverse illumination conditions.
To alleviate this problem, we propose an instance-aware modality-balanced optimization strategy. To be specific, we additionally preserve visible and thermal decoder branches besides the fusion branch during the training phase. 
Based on the trident structure of the decoder, as shown in Fig.~\ref{fig:decoder}, we can then obtain three sets of predicted slots. 
Finally, an instance-wise dynamic loss is devised to adaptively adjust the contributions of each pedestrian instance to the model learning.
     
The main contributions of the paper are summarized as follows:
\begin{itemize}
    \item We propose an end-to-end multispectral pedestrian detection model, dubbed as MS-DETR, which extends DETR from single-modal detection task into multispectral case by introducing a multi-modal Transformer decoder. 
    \item  A novel loosely coupled fusion strategy is proposed to improve the feature fusion efficiency towards multi-modal detection task. It shows great potential in DETR-like multi-modal framework compared with other fusion strategies.
    \item An instance-aware modality-balanced optimization strategy is designed to align three sets of slots produced by three detection branches, i.e., visible, thermal and fusion branches. Based on these aligned slots, we use a dynamic loss to measure and adjust the contribution of each instance. 
    \item  We conduct thorough experiments to verify the effectiveness of the proposed method on three widely used datasets, namely KAIST, CVC-14, and LLVIP. Experimental results demonstrate the superiority of the proposed MS-DETR.
\end{itemize}

\section{Related Works}
\subsection{Multispectral Pedestrian Detection}
Since the release of large-scale multi-modal datasets like KAIST Multispectral Pedestrian Benchmark~\cite{KAIST}, CVC14 datasets~\cite{CVC14Dataset}, and LLVIP~\cite{jia2021llvip} etc, a variety of multispectral pedestrian detection methods were proposed~\cite{KAIST,choi2016multi,konig2017fully,liu2016multispectral,park2018unified,li2019illumination,guan2019fusion,limultispectral,cao2019box,zhang2021guided, dasgupta2022spatio, zhu2023multi,li2022confidence,zhang2019weakly,zhou2020improving,kim2021mlpd,zhang2019cross,zhang2020multispectral,xu2017learning,zhang2021deep,liu2021deep,kim2021uncertainty}. Hwang \etal~\cite{KAIST} extended the aggregated channel features (ACF) by incorporating the intensity and histogram of oriented gradient (HOG) features of the thermal
channel to propose an ACF+T+THOG method. However, ACF+T+THOG features lack discrimination in challenging circumstances, such as in cases of tiny appearance and poor visibility, therefore, Choi~\etal~\cite{choi2016multi} proposed to learn discriminative features through CNN and support vector regression (SVR). It is critical for multispectral pedestrian detectors to fuse multi-modal features. K\"onig~\etal~\cite{konig2017fully} developed a multispectral region proposal network (RPN) based on Faster R-CNN~\cite{ren2015faster}. It fused the visible and thermal features by concatenation, and concluded that the best detection performance was obtained through halfway fusion~\cite{liu2016multispectral}. 

Apart from concatenation, there are other fusion strategies~\cite{chen2023grid,chen2023temporal}. 
Park~\etal~\cite{park2018unified} estimated the detection probabilities of different modalities by a channel weighting fusion layer, and then utilized an accumulated probability fusion layer to combine them. Some works incorporated illumination information of images as prior knowledge, and introduced illumination-aware networks to adaptively merge visible and thermal features~\cite{li2019illumination,guan2019fusion,liu2021deep}. Guan~\etal~\cite{guan2019fusion} developed a two-stream deep neural networks, which not only utilized two illumination-aware networks to produce fusion weights, but also trained the detector by jointly learning pedestrian detection and semantic segmentation tasks. Actually, many works~\cite{limultispectral,liu2021deep,guan2019fusion,cao2019box} used semantic segmentation as an auxiliary task, which was proved to be an effective way to boost the performance of multispectral pedestrian detection. But the researchers should first obtain the semantic segmentation labels manually. Except for methods who made use of illumination information, Kim~\etal~\cite{kim2021uncertainty} approached an uncertainty-aware multispectral pedestrian detection framework, which considered two types of uncertainties and fused the multispectral features by an uncertainty-aware feature fusion module. 
To well fuse images, Zhang~\etal~\cite{zhang2021guided} converted detection labels into elliptical mask labels to roughly match the shape of pedestrians.
Li~\etal~\cite{li2022confidence} used the dense fusion strategy to extract multilevel multispectral representations and then combined the prediction results of different modalities based on Dempster's rule of combination. It not only integrated multi-modal information~\cite{chen2022global,chen2023local,dasgupta2022spatio}, but also provided a reliable prediction. 
In a word, 
existing methods generally incorporate various feature fusion modules within the model, and some of them 
introduce additional prior knowledge to further improve the detection results.

Recently, some researchers emphasized the misaligment problem of KAIST~\cite{KAIST} and CVC-14~\cite{CVC14Dataset}, and proposed some strategies to reduce the influence of modality-misalingment~\cite{zhang2019weakly,zhou2020improving,kim2021mlpd}. Zhang~\etal~\cite{zhang2019weakly} directly designed a Region Feature Alignment (RFA) module to adaptively compensate misalignment of feature maps in two modalities. Zhou~\etal~\cite{zhou2020improving} developed an illumination-aware feature alignment module to select complementary features according to illumination conditions,
which obtained better detection results. Aiming at the unpaired multispectral pedestrian detection issue, Kim~\etal~\cite{kim2021mlpd} leveraged multi-label learning to extract the state-aware features. Zhu~\etal~\cite{zhu2023multi} used multi-modal feature pyramid Transformer to overcome the large visual differences and misalignment between two modalities. Nonetheless, the misalignment in multi-modal images remains a difficult issue to solve. 


All above mentioned deep-learning based methods usually
require explicit priors or the post-processing steps. In this
paper, we propose an end-to-end multispectral pedestrian
detection model without the need of any anchor mechanism,
post-processing steps or the illumination/uncertainty prior, and
is also robust to the misalignment of modalities.

\begin{figure}[t]
  \centering
  \includegraphics[scale=0.45]{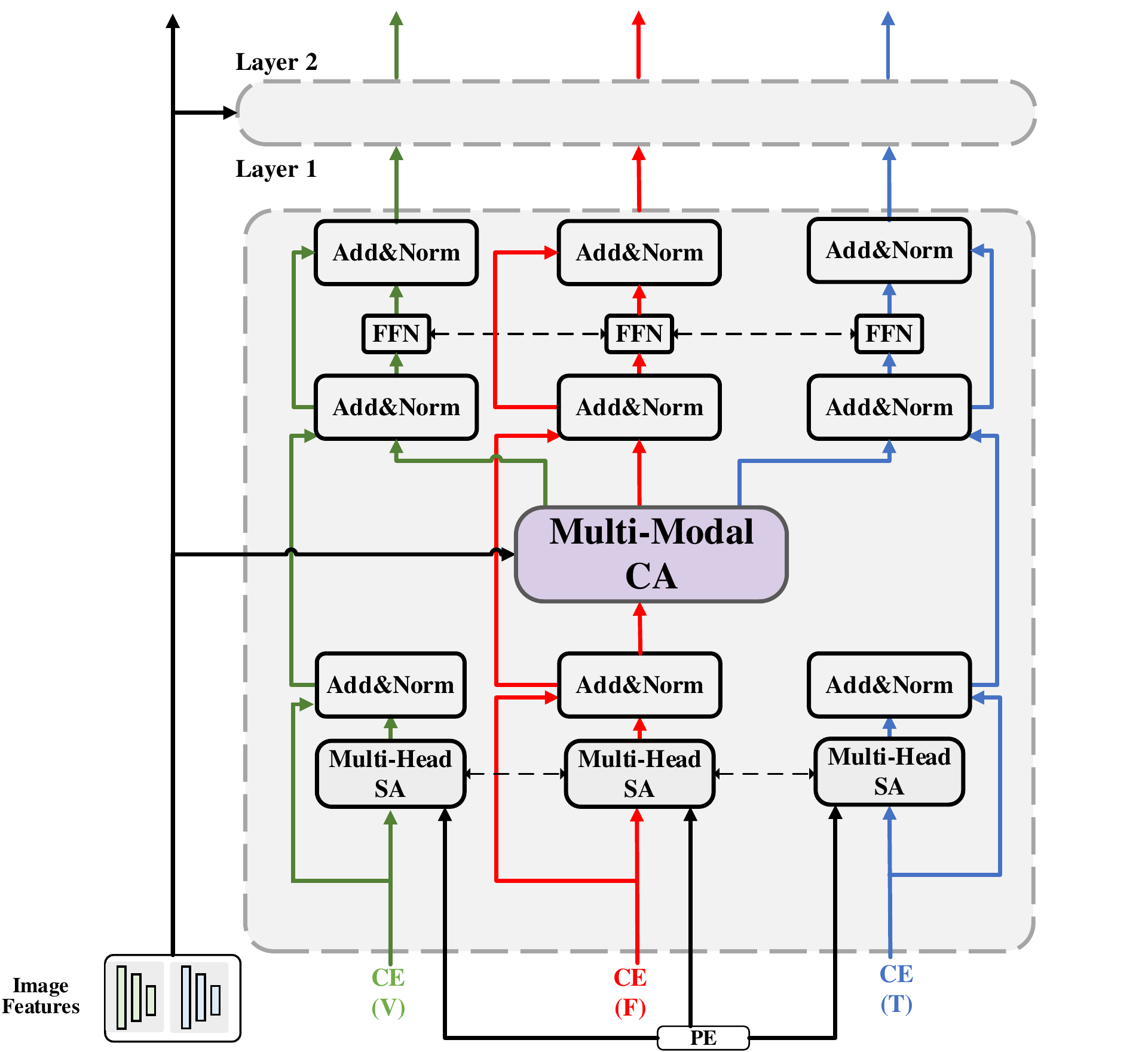}
  \caption{The multi-modal Transformer decoder of MS-DETR. It has three branches, \ie visible (V), fusion (F), and thermal (T), where PE, CE, SA, CA and FFN are acronyms for Positional Encodings, Content Embeddings, Self-Attention, Cross-Attention and feed forward network. Dashed lines indicate shared parameters.}
  \label{fig:decoder}
\end{figure}

\subsection{Object Detection with Transformer}
ViT-FRCNN~\cite{beal2020toward} is the first attempt to use a pre-trained vision Transformer (ViT)~\cite{dosovitskiy2020image} as the backbone for a Faster R-CNN~\cite{ren2015faster} object detector. However, the design cannot get rid of the reliance on CNNs. You Only Look at One Sequence (YOLOS)~\cite{fang2021you} performs 2D object-level and region-level recognition based on the vanilla ViT with the fewest possible modifications, region priors, as well as inductive biases of the object detection task, demonstrating that 2D object detection can be accomplished in a pure sequence-to-sequence manner by taking a sequence of fixed-sized non-overlapping image patches as input.
DETR~\cite{DETR}, as one of the breakthroughs in object detection, treats object detection as a direct set prediction task without any hand-crafted prior knowledge like anchor generation or NMS. It is trained in an end-to-end manner with a set-based loss function, which performs bipartite matching between the predicted and the groundtruth bounding box~\cite{bulat2022fs}. DETR greatly simplifies the pipeline of object detection, and guides the tendency of object detection. Nevertheless, it suffers from slow training convergence. Several works have been proposed to improve it~\cite{meng2021conditionalDETR,dai2021upDETR,dai2021dynamicDETR,zhu2020deformableDETR,zhang2022SAMDETR,liu2021dabDETR}. Conditional DETR~\cite{meng2021conditionalDETR} presented a conditional cross-attention mechanism to accelerate DETR's training, and it converged 6.7$\times$ faster than original DETR. Deformable DETR~\cite{zhu2020deformableDETR} was proposed to make the attention modules
only attend to a small set of key sampling points around a reference, which makes it easy to aggregate multi-scale features. DAB-DETR~\cite{liu2021dabDETR} directly used box coordinates as queries to improve the query-to-feature similarity and to eliminate slow training convergence issue in DETR. Dynamic DETR~\cite{dai2021dynamicDETR} developed an alternative solution to boost the training convergence of DETR by replacing the cross-attention module in DETR decoder with an ROI-based dynamic attention.  

Apart from the single-modal detectors, there are also some multi-modal detectors build upon Transformer. Kamath \etal~\cite{kamath2021mdetr} proposed an modulated detection model, MDETR, for end-to-end multi-modal understanding, where the visual and text features extracted from a convolutional backbone and a language model were concatenated and passed to the DETR~\cite{DETR} model for detection. However, the flattening operation on the visual features 
would destroy the spatial structure of images. Therefore, Maaz \etal~\cite{maaz2022class} took the late multi-modal fusion strategy into MDETR framework and also introduced the multi-scale spatial context deformable attention to boost the training.
MT-DETR~\cite{chu2023mt}, an end-to-end multi-modal detection model, used the residual fusion module and confident fusion module to fuse camera, lidar, radar, and time features for autonomous driving. They are presented for vision-language or camera-lidar alignment-related tasks, which are different from multispectral pedestrian detection that the inputs are both images. Zhu~\etal~\cite{zhu2023multi} designed intra-modal feature pyramid Transformer and inter-modal feature pyramid Transformer to improve the semantic representations of features and make the model insensitive to misaligned images. In this paper, we propose to solve the problem of misalignment from the fusion aspect, and further consider the modality imbalance problem in multispectral pedestrian detection. 
\begin{figure*}[t]
  \centering
  \includegraphics[scale=1]{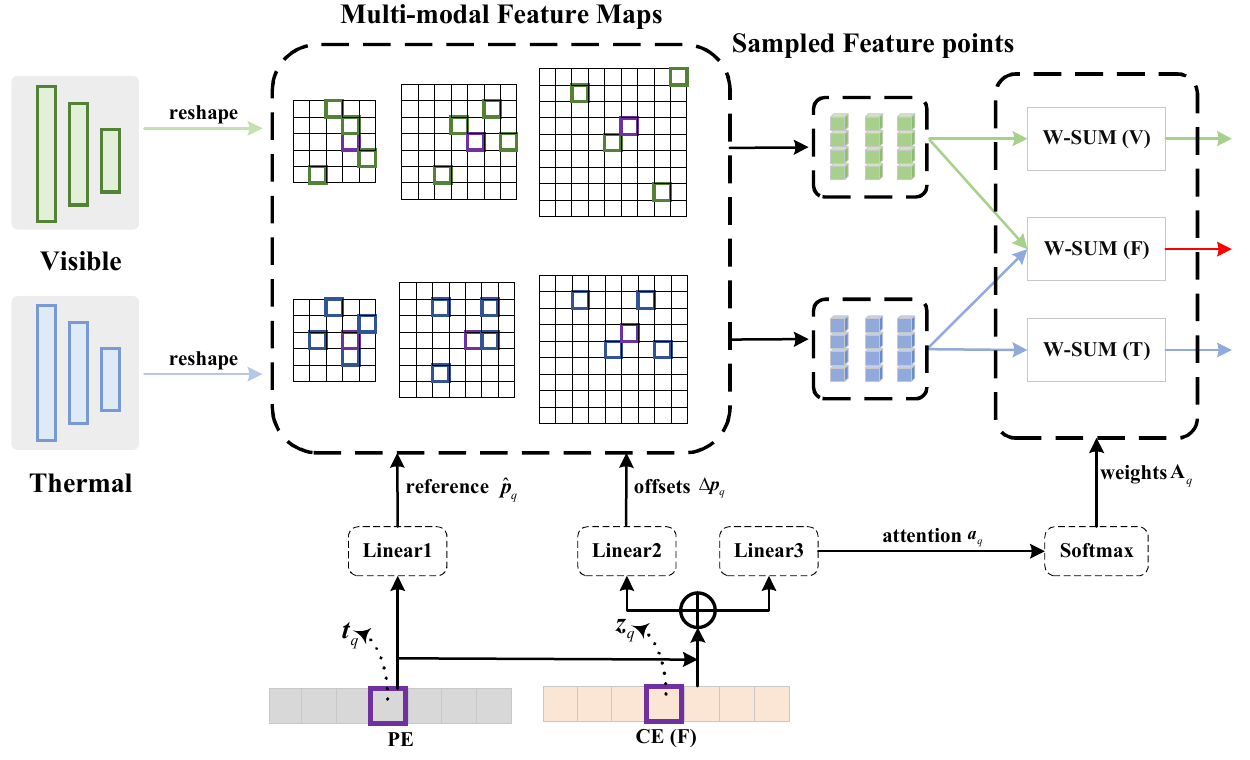}
  \caption{Details of multi-modal cross-attention (CA) module. The positional encodings pass through a linear layer to predict reference points, and they are then combined with content embeddings of fusion branch to predict offsets and their corresponding attention weights. Given a group of reference points and offsets, two groups of feature points are sampled from multi-modal multi-scale features. These sampled feature points are fused by weighted sum operations. }
  \label{fig:fusion}
\end{figure*}

\section{Proposed Method}
\subsection{Overview}
We propose MS-DETR, an end-to-end multispectral pedestrian detection model. As shown in Fig.~\ref{fig:overall_architecture}, the proposed model is composed of three parts: two modality-specific CNN backbones, two modality-specific Transformer encoders and a multi-modal Transformer decoder. 

Given a pair of visible image $\mathbf{I}^{\rm{V}}$ and thermal image $\mathbf{I}^{\rm{T}}$, each modality-specific CNN backbone first extracts multi-scale feature maps. Then, these feature maps are fed into corresponding modality-specific encoder to aggregate the modality-specific multi-scale features through deformable self-attention. These two steps can be formulated as
\begin{equation}
  \{\mathbf{E}^m_l\}_{l=1}^L = \phi^m(\varphi^m(\mathbf{I}^m)), m \in \{\rm{V},\rm{T}\},
  \label{eq:1}
\end{equation}
where $\varphi^m(\cdot)$ and $\phi^m(\cdot)$ denote the CNN backbone and Transformer encoder of modality $m$. $\{\mathbf{E}^m_l\}_{l=1}^L$ represents the refined multi-scale feature maps, and $l$ is the index of scale.

The subsequent multi-modal Transformer decoder takes the multi-scale feature maps $\{\mathbf{E}^m_l\}_{l=1}^L$ of both modalities as the inputs. Besides, the learnable content embeddings (CE) combined with positional encodings (PE) are input to the decoder to perform cross-attention with the encoded input features. 
As shown in Fig.~\ref{fig:decoder}, instead of keeping the fusion branch alone, we preserve the visible and thermal branch additionally in the decoder. With the help of the constituted trident architecture, three sets of predicted slots can be obtained through the visible, thermal and fusion branches of the decoder, respectively.

To alleviate the modality-imbalance problem as stated previously, an instance-aware modality-balanced strategy is introduced to optimize the model. Specifically, an instance-wise dynamic weighting loss is enforced on top of the decoder to adaptively adjust the contributions of two modalities for each pedestrian instance.

Since our CNN backbones and modality-specific Transformer encoders are similar to previous works, we will not repeat them. Interested readers can refer to \cite{zhu2020deformableDETR} for more details. 
In the following, we will elaborate the proposed multi-modal Transformer decoder and the instance-aware modality-balanced optimization strategy.

\subsection{Multi-Modal Transformer Decoder}
The detailed architecture of multi-modal Transformer decoder is shown in Fig.~\ref{fig:decoder}. Our multi-modal Transformer decoder has three branches, i.e., visible (V), fusion (F) and thermal (T) branches. Correspondingly, we have three content embeddings (CEs) and a shared positional encoding (PE). In each layer, the multi-head Self-Attention (SA) and Feed-Forward Network (FFN) of visible, fusion and thermal branches have the same parameters, and all the three branches share a common multi-modal cross-attention (CA) module. 
The proposed loosely coupled fusion is achieved in the multi-modal cross-attention module. In the following, we will elaborate this module.

\begin{table*}[t!]
    \caption{Comparisons with the state-of-the-art methods on the KAIST dataset}
    \centering
    \setlength{\tabcolsep}{7mm}{
        \begin{tabular}{c|ccc|ccc}
            \hline
            \multirow{3}{*}{Methods}                   & \multicolumn{6}{c}{$MR^{-2}$ (IoU=0.5)}                                             \\ \cline{2-7}
                                                       & \multicolumn{3}{c}{\textit{Reasonable}} & \multicolumn{3}{|c}{\textit{All}}         \\ \cline{2-7} 
                                                       & {R-All} & {R-Day} & {R-Night} & {All}   & {Day}   & {Night}                         \\  \hline
            ACF~\cite{KAIST}                           & {47.32} & {42.57} & {56.17}   & {67.74} & {64.32} & {75.06}                         \\
            Halfway Fusion~\cite{liu2016multispectral} & {25.75} & {24.88} & {26.59}   & {49.20} & {47.67} & {52.35}                         \\
            Fusion RPN+BF~\cite{konig2017fully}        & {18.29} & {19.57} & {16.27}   & {51.67} & {52.30} & {51.09}                         \\
            IAF R-CNN~\cite{li2019illumination}        & {15.73} & {14.55} & {18.26}   & {44.24} & {42.47} & {47.70}                         \\
            IATDNN+IASS~\cite{guan2019fusion}          & {14.95} & {14.67} & {15.72}   & {46.45} & {46.77} & {46.34}                         \\
            CIAN~\cite{zhang2019cross}                 & {14.12} & {14.78} & {11.13}   & {35.57} & {36.06} & {32.38}                         \\
            MSDS-RCNN~\cite{limultispectral}           & {11.34} & {10.54} & {12.94}   & {34.20} & {32.12} & {38.83}                         \\
            AR-CNN~\cite{zhang2019weakly}              & {9.34}  & {9.94}  & {8.38}    & {34.95} & {34.36} & {36.12}                         \\ 
            ProbEn$_3$~\cite{chen2022multimodal}              & 9.07  & 7.66  & 4.89    & - & - & -  \\
            MBNet~\cite{zhou2020improving}             & {8.13}  & {8.28}  & {7.86}    & {31.87} & {32.39} & {30.95}                         \\
            CMPD~\cite{li2022confidence}               & {8.16}  & {8.77}  & {7.31}    & {28.98} & \underline{28.30} & {30.56}
            \\
            Faster RCNN+MFPT~\cite{zhu2023multi}         & {7.72}  & {8.26}  & {4.53}    & {-}     & {-}     & {-}                             \\
            MLPD~\cite{kim2021mlpd}                    & {7.58}  & {7.96}  & {6.95}    & {28.49} & {28.39} & {28.69}                         \\
            AANet-Faster RCNN~\cite{aanet-2023}        & 6.91  & \textbf{6.66}     & 7.31    & -   & - & - \\
            MCHE-CF~\cite{mch-li2023}                  & 6.71  & \underline{7.58}    & 5.52  & -    & -   & -\\
            GAFF~\cite{zhang2021guided}                & \underline{6.48}  & 8.35  & \underline{3.46}    & \underline{27.30} & {30.59} & \underline{19.22}\\ \hline
            MS-DETR                                    & {\textbf{6.13}}  & {7.78} & {\textbf{3.18}}   & {\textbf{20.64}}  & {\textbf{22.76}} & {\textbf{15.77}}  \\ \hline
        \end{tabular}
    }
    \begin{tablenotes}
            \item ``-" indicates the result is not provided.
    \end{tablenotes}
    \label{tab:KAIST}
\end{table*}

\textbf{Loosely coupled fusion in multi-modal CA module.}
The multi-modal cross-attention builds upon deformable attention, where only a small set of key sampling points around a reference
point are considered, regardless of the spatial size of the feature maps.
Fig.~\ref{fig:fusion} shows the detailed architecture of multi-modal cross-attention module. Specifically, the $q$-th positional encoding $\boldsymbol{t}_{q}$ is taken into $\operatorname{Linear1}$ to obtain the reference point's coordinate $\hat{\boldsymbol{p}}_q \in \mathbb{R}^{2}$,
\begin{equation}
\begin{split}
\hat{\boldsymbol{p}_q} &=\operatorname{Linear1}(\boldsymbol{t}_{q}).
\end{split}
\label{eq:4}
\end{equation}
Meanwhile, the $q$-th positional encoding $\boldsymbol{t}_{q}$ is also combined with content embedding $\boldsymbol{z}_{q}$ of fusion branch to directly predict offsets of sampling points $\Delta \boldsymbol{p}_{q} \in \mathbb{R}^{|m| \times H \times L \times K \times 2}$ and their corresponding attention weights $\boldsymbol{a}_{q} \in \mathbb{R}^{|m| \times H \times L \times K}$ through two independent linear layers, 
\begin{equation}
\begin{split}
\Delta \boldsymbol{p}_{q} &=\operatorname{Linear2}(\boldsymbol{z}_{q} + \boldsymbol{t}_{q}), \\
\boldsymbol{a}_{q} &= \operatorname{Linear3}(\boldsymbol{z}_{q} + \boldsymbol{t}_{q}),
\end{split}
\label{eq:4}
\end{equation}
where $m=\{\rm{V},\rm{T}\}$ indicates the visible and thermal modalities, and $|m|$ is the number of modalities. $H$, $L$ and $K$ are the number of attention heads, features scales and the sampling points in a certain feature map.

After obtaining the prominent key elements and their corresponding attention weights of different feature scales and modalities, we use the softmax function to normalize these weights, \ie, $\boldsymbol{A}_{q} = \operatorname{Softmax}(\boldsymbol{a}_{q})$. The loosely coupled fusion is conducted only on these sampled points and their normalized attention weights of different modalities. The fused features can be obtained by
\begin{equation}
\begin{split}
&\operatorname{Multi-Modal CA}_{\rm{F}} = \\ & \sum_{h=1}^H \boldsymbol{W}_h\left[\sum_{m}\sum_{l=1}^L \sum_{k=1}^K A_{q m h l k} \cdot \boldsymbol{W}_h^{\prime}\boldsymbol{E}_l^m\left(\alpha_l\left(\hat{\boldsymbol{p}}_q\right)+\Delta \boldsymbol{p}_{q m h l k}\right)\right],
\end{split}
\label{eq:5}
\end{equation}
where $\hat{\boldsymbol{p}}_q$ is the reference point of $q$-th query. $m$ indicates the modality. $h$, $l$, and $k$ represent the attention head, feature scale and sampling point, respectively. $\alpha_l\left(\hat{\boldsymbol{p}}_q\right)$ re-scales $\hat{\boldsymbol{p}}_q$ to the input feature maps of the $l$-th scale. $\Delta \boldsymbol{p}_{q h l k}$ and $A_{q h l k}$ represent the sampling offset and the scalar attention weight of the $k$-th sampling point in the $l$-th feature level for the $h$-th attention head.
We can observe from (\ref{eq:5}) that the fusion of visible and thermal features is only conducted during the process of aggregating prominent key elements of two modalities, thus we call it ``loosely coupled fusion". 

Based on the ``loosely coupled fusion", we can easily introduce extra two detection branches to avoid under-optimizing uni-modal representation~\cite{peng2022balanced,wu2022characterizing}.  
As shown in Fig.~\ref{fig:fusion}, we divide the offsets $\Delta \boldsymbol{p}_{q}$ into $\Delta \boldsymbol{p}_{q}^{\rm{V}}$ and $\Delta \boldsymbol{p}_{q}^{\rm{T}}$. Correspondingly, the attention weights $\boldsymbol{a}_{q}$ are also splitted to two modality specific components $\boldsymbol{a}_{q}^{\rm{V}}$ and $\boldsymbol{a}_{q}^{\rm{T}}$. After the normalization of softmax, we obtain the modality specific attention weights $\boldsymbol{A}_{q}^{\rm{V}} = \operatorname{Softmax}(\boldsymbol{a}_{q}^{\rm{V}})$ and $\boldsymbol{A}_{q}^{\rm{T}} = \operatorname{Softmax}(\boldsymbol{a}_{q}^{\rm{T}})$. Therefore, extra two groups of uni-modal features in the multi-modal cross-attention module are formulated by
\begin{equation}
\begin{split}
&\operatorname{Multi-Modal CA}_{\rm{V}} = \\ & \sum_{h=1}^H \boldsymbol{W}_h\left[\sum_{l=1}^L \sum_{k=1}^K A_{q h l k}^{\rm{V}} \cdot \boldsymbol{W}_h^{\prime}\boldsymbol{E}_l^{\rm{V}}\left(\alpha_l\left(\hat{\boldsymbol{p}}_q\right)+\Delta \boldsymbol{p}_{q h l k}^{\rm{V}}\right)\right],
\end{split}
\label{eq:V}
\end{equation}

\begin{equation}
\setlength{\abovedisplayskip}{3pt}
\setlength{\belowdisplayskip}{3pt}
\begin{split}
&\operatorname{Multi-Modal CA}_{\rm{T}} = \\ & \sum_{h=1}^H \boldsymbol{W}_h\left[\sum_{l=1}^L \sum_{k=1}^K A_{q h l k}^{\rm{T}} \cdot \boldsymbol{W}_h^{\prime}\boldsymbol{E}_l^{\rm{T}}\left(\alpha_l\left(\hat{\boldsymbol{p}}_q\right)+\Delta \boldsymbol{p}_{q h l k}^{\rm{T}}\right)\right].
\end{split}
\label{eq:T}
\end{equation}

\subsection{Instance-aware Modality-Balanced Optimization}
By passing the decoder output to the modality-specific detection head, we can obtain three sets of predictions. Let $i\in \{\rm{V},\rm{T},\rm{F}\}$ indicates different branches, these predictions can be represented by $\{\hat{y}^{i}_n\}_{n=1}^{N}$, $i \in \{\rm{V},\rm{T},\rm{F}\}$. As the distinct characteristics of visible and thermal images, their predictions may not be consistent. To balance the contribution of different modalities, we compare and then select the optimal permutation of slots predictions in current iteration. After that, the predictions of all the three branches are re-arranged according to the selected permutation $\hat{\sigma}$.

To obtain $\hat{\sigma}$, we should firstly calculate the optimal permutation for each branch $i$ 
\begin{equation}
\setlength{\abovedisplayskip}{3pt}
\setlength{\belowdisplayskip}{3pt}
\hat{\sigma}^{i}=\underset{\sigma \in \mathfrak{S}_N}{\arg \min } \sum_{n=1}^{N} \mathcal{L}_{match}\left(y_n, \hat{y}_{\sigma(n)}^{i}\right), i\in\{\rm{V},\rm{T},\rm{F}\},
\label{eq:6}
\end{equation}
where $\mathfrak{S}_N$ is the set of all permutations of $N$ elements. $\mathcal{L}_{match}\left(y_n, \hat{y}_{\sigma(n)}^{i}\right)$ is the matching cost between ground-truth $y_n$ and the prediction produced by $\sigma(n)$-th anchor box of modality $i$, and can be computed by 
\begin{equation}
\setlength{\abovedisplayskip}{3pt}
\setlength{\belowdisplayskip}{3pt}
\begin{split}
\mathcal{L}_{match}\left(y_n, \hat{y}_{\sigma(n)}^{i}\right)  = & \mathcal{L}_{focal}\left(y_n, \hat{y}_{\sigma(n)}^{i}\right)  + \mathcal{L}_{L1}\left(y_n, \hat{y}_{\sigma(n)}^{i}\right) \\&+ \mathcal{L}_{GIoU}\left(y_n, \hat{y}_{\sigma(n)}^{i}\right).
\label{eq:7}
\end{split}
\end{equation}
By comparing the matching costs under the permutations $\hat{\sigma}^{\rm{V}}$, $\hat{\sigma}^{\rm{T}}$ and $\hat{\sigma}^{\rm{F}}$, we obtain the optimal permutation
\begin{equation}
\setlength{\abovedisplayskip}{3pt}
\setlength{\belowdisplayskip}{3pt}
\hat{\sigma} =\underset{\sigma \in \{\hat{\sigma}^{i}\}}{\arg \min } \sum_{n=1}^{N} \mathcal{L}_{match}\left(y_n, \hat{y}_{\sigma(n)}^{i}\right), i\in\{\rm{V},\rm{T},\rm{F}\}.
\label{eq:8}
\end{equation}
Then, the slots of other branches are re-arranged according to $\hat{\sigma}$, and their matching costs are updated under the new permutation.  

In order to further boost the detection performance and to balance the training, we propose an instance-wise dynamic loss. Specifically, the matching costs are calculated under the optimal permutation in instance-level, and then, they are utilized to measure the degree of prediction confidence for different instances to adaptively adjust the instance-level weights of three branch. Suppose that there are $T$ pedestrian instances, we use $c_{\hat{\sigma}(j)}^i$ to represent the instance-level matching cost of $j$-th instance in modality $i$ under the optimal permutation $\hat{\sigma}$. The dynamic fusion parameters $\lambda_{\sigma(j)}^i$, $\left(i \in \{\rm{V},\rm{T},\rm{F}\}\right)$ are obtained through
\begin{equation}
\setlength{\abovedisplayskip}{3pt}
\setlength{\belowdisplayskip}{3pt}
\lambda_{\sigma(j)}^i = \frac{e^{c_{\sigma(j)}^i}}{e^{c_{\sigma(j)}^{\rm{V}}} + e^{c_{\sigma(j)}^{\rm{F}}} + e^{c_{\sigma(j)}^{\rm{T}}}}, j=1,\cdots,T.
\label{eq:8}
\end{equation}

Then the loss of modality $i$ is composed of a dynamic loss term for predicted pedestrians and a focal loss term for predicted non-pedestrians:
\begin{equation}
\setlength{\abovedisplayskip}{3pt}
\setlength{\belowdisplayskip}{3pt}
\mathcal{L}^i = \sum_{s=1}^{T} \lambda_{\hat{\sigma}(s)}^i \cdot \mathcal{L}_{match}\left(y_s, \hat{y}_{\hat{\sigma}(s)}^i\right) + \sum_{s=T+1}^N \mathcal{L}_{focal}\left(y_s, \hat{y}_{\hat{\sigma}(s)}^i\right).
\label{eq:9}
\end{equation}

Finally, the total loss of our MS-DETR is
\begin{equation}
\setlength{\abovedisplayskip}{3pt}
\setlength{\belowdisplayskip}{3pt}
\mathcal{L} = \mathcal{L}^{\rm{F}} + \mathcal{L}^{\rm{V}} + \mathcal{L}^{\rm{T}}.
\label{eq:10}
\end{equation}

\subsection{Inference process}
In the training stage, we preserve visible, thermal and fusion branches to calculate the instance-aware loss function for  modality balance. After modality-balanced optimization, any of the three branches has the ability to obtain accurate detection results, which has been verified, and the results can be found in Table~\ref{tab:balance}. Therefore, during inference, we can use one of the three branches to obtain results. And we use the results of fusion branch by default. 

\begin{table*}[t]
    \caption{Performance comparisons on the six subsets of KAIST dataset}
    \centering
    \setlength{\tabcolsep}{6mm}{
        \begin{tabular}{c|cccccc|c}
            \hline
            Methods                                    & Near  & Medium & Far   & None  & Partial & Heavy & {All} \\ \hline
            ACF~\cite{KAIST}                           & 28.74 & 53.67  & 88.20 & 62.94 & 81.40   & 88.08 & 67.74 \\
            Halfway Fusion~\cite{liu2016multispectral} & 8.13  & 30.34  & 75.70 & 43.13 & 65.21   & 74.36 & 49.20 \\
            Fusion RPN+BF~\cite{konig2017fully}        & \underline{0.04}  & 30.87  & 88.86 & 47.45 & 56.10   & 72.20 & 51.67 \\
            IAF R-CNN~\cite{li2019illumination}        & 0.96  & 25.54  & 77.84 & 40.17 & 48.40   & 69.76 & 44.24 \\
            IATDNN+IASS~\cite{guan2019fusion}          & \underline{0.04}  & 28.55  & 83.42 & 45.43 & 46.25   & 64.57 & 46.45 \\
            CIAN~\cite{zhang2019cross}                 & 3.71  & 19.04  & 55.82 & 30.31 & 41.57   & 62.48 & 35.57 \\
            MSDS-RCNN~\cite{limultispectral}           & 1.29  & 16.18  & 63.76 & 29.94 & 38.46   & 63.30 & 34.20 \\
            AR-CNN~\cite{zhang2019weakly}              & \textbf{0.00}  & 16.08  & 69.00 & 31.40 & 38.63   & 55.73 & 34.95 \\
            MBNet~\cite{zhou2020improving}             & \textbf{0.00}  & 16.07  & 55.99 & 27.74 & 35.43   & 59.14 & 31.87 \\
            CMPD~\cite{li2022confidence}               & \textbf{0.00}  & 12.99  & 51.22 & 24.04 & 33.88   & 59.37 & 28.98 \\
            MLPD~\cite{kim2021mlpd}                    & \textbf{0.00}  & 12.10  & 52.79 & 25.18 & 29.84   & 55.05 & 28.49 \\
            MCHE-CF~\cite{mch-li2023}                   & \textbf{0.00}  & \underline{10.40}  & \underline{36.75} & \underline{18.05} & 24.33   & 54.43 & - \\
            GAFF~\cite{zhang2021guided}                & \textbf{0.00}  & 13.23  & 46.87 & 23.83 & \underline{24.31}   & \textbf{47.97} & \underline{27.30} \\ \hline
            MS-DETR                                    & \textbf{0.00}  & \textbf{9.70}   & \textbf{32.41} & \textbf{16.54} & \textbf{23.61}   & \underline{48.71} & \textbf{20.64} \\ \hline 
        \end{tabular}
    }
    \label{tab:KAISTsubset}
\end{table*}

\section{Experiments}
\subsection{Datasets and Evaluation Metric}
Following previous methods, we conduct experiments on three popular datasets, i.e. KAIST, CVC-14 and LLVIP.

\textbf{KAIST.} Originally, the KAIST Multispectral Pedestrian Benchmark~\cite{KAIST} contains 95,328 color-thermal image pairs with 103,128 dense annotations covering 1,182 unique pedestrians. For fair comparison with previous methods, we use 7,601 images with sanitized annotations ~\cite{limultispectral} for training and 2,252 images with improved annotations by Liu~\etal~\cite{liu2016multispectral} for testing. Specifically, in the test set, 1,455 images are captured during daytime and the remaining during nighttime.

\begin{figure}[t]
  \centering
  \includegraphics[width=0.4\textwidth]{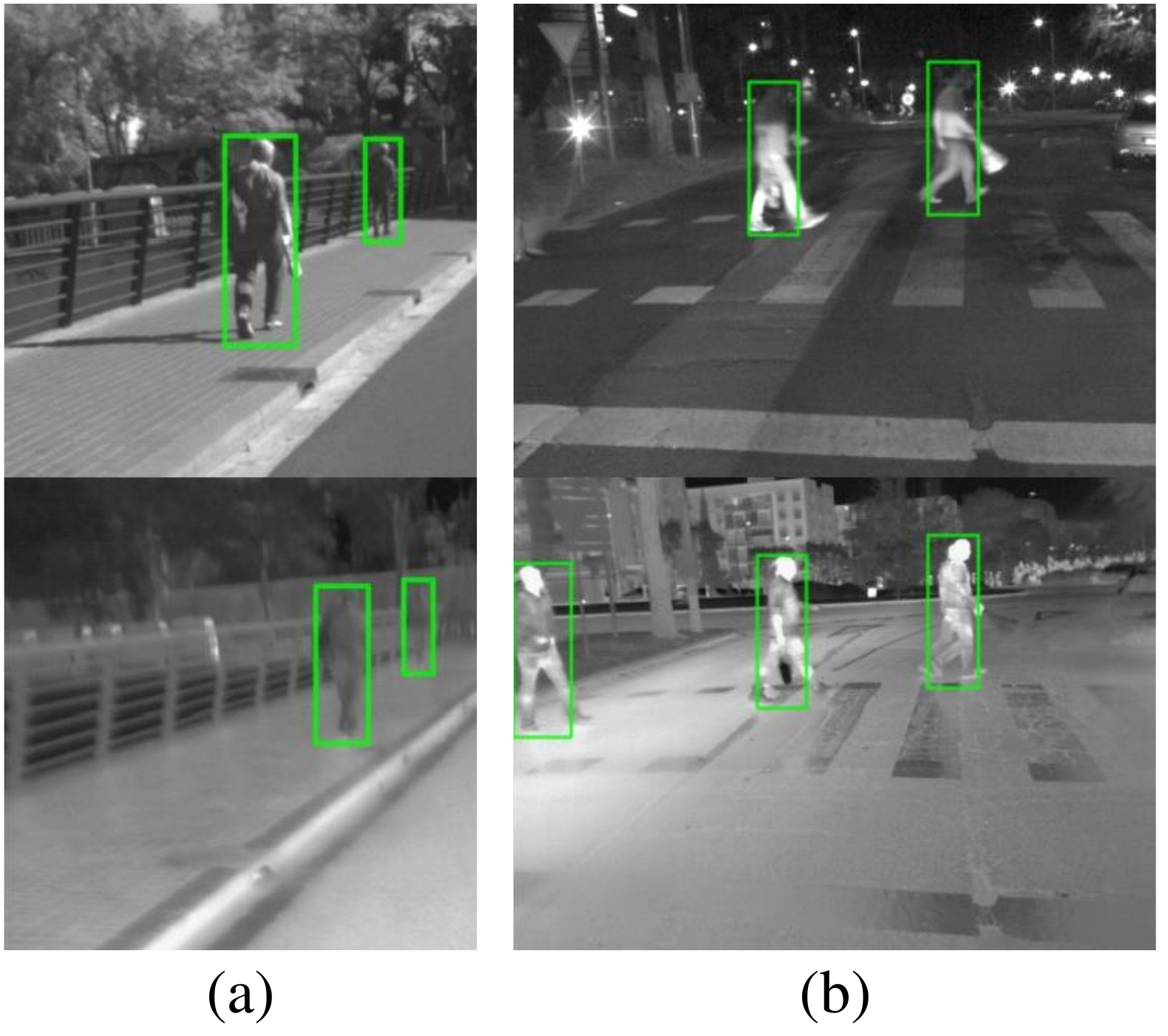}
  \caption{The visualization examples of position shift problem in CVC-14 dataset, where the pedestrians in (a) are grossly misaligned in
spatial dimensions and the number of pedestrians in (b) are unpaired.}
  \label{fig:cvc14_misalign}
\end{figure}

\textbf{CVC-14.} The CVC-14 dataset~\cite{CVC14Dataset} has 7,085 visible (grayscale) and thermal image pairs for training, where 3,695 pairs are captured during daytime and 3,390 pairs during nighttime. The test set is composed of 1,433 frames. 
Since the cameras are not well calibrated, the annotations in CVC-14 are individually provided in each modality. Besides, as shown in Fig.~\ref{fig:cvc14_misalign}, the CVC-14 dataset suffers from a severer position shift problem, which makes it more difficult.

\textbf{LLVIP.} LLVIP~\cite{jia2021llvip} is a paired visible-infrared dataset for low-light vision. It has 15,488 pairs of aligned images, with 42,437 annotated pedestrians totally. Most of them are taken under the dark scenario by surveillance cameras. The quality of original visible and thermal images in LLVIP dataset is significantly higher than that of the KAIST and CVC-14 datasets. The visible and thermal images have the resolution of $1920 \times 1080$ and $1280 \times 720$ respectively.

\textbf{Evaluation metric.}
For KAIST and CVC-14, we employ the log miss rate ($MR^{-2}$) averaged over the false positive per image (FPPI) with a range of $[10^{-2},10^0]$ for evaluation~\cite{dollar2011pedestrian,KAIST}. A lower $MR^{-2}$ indicates better performance. Moreover, we will report the performance under two evaluation settings for KAIST dataset, \ie ``\textit{Reasonable}'' and ``\textit{All}". The former only considers the pedestrians taller than 55 pixels with no or partial occlusions while the latter uses whole test data including small or heavy occluded pedestrians. Therefore, the ``\textit{All}" setting is more challenging than ``\textit{Reasonable}" setting.

For LLVIP, we employ the commonly used average precision metrics ($AP_{iou}$) under different IoU thresholds as evaluation metrics. Specifically, we select $AP_{0.5}$, $AP_{0.75}$, and $AP$ as three metrics for comparison. The $AP$ metric represents the mean $AP_{iou}$, whose IoU ranging from 0.50 to 0.95 with a stride of 0.05. 

\begin{figure*}[t]
  \centering
  \includegraphics[width=0.9\textwidth]{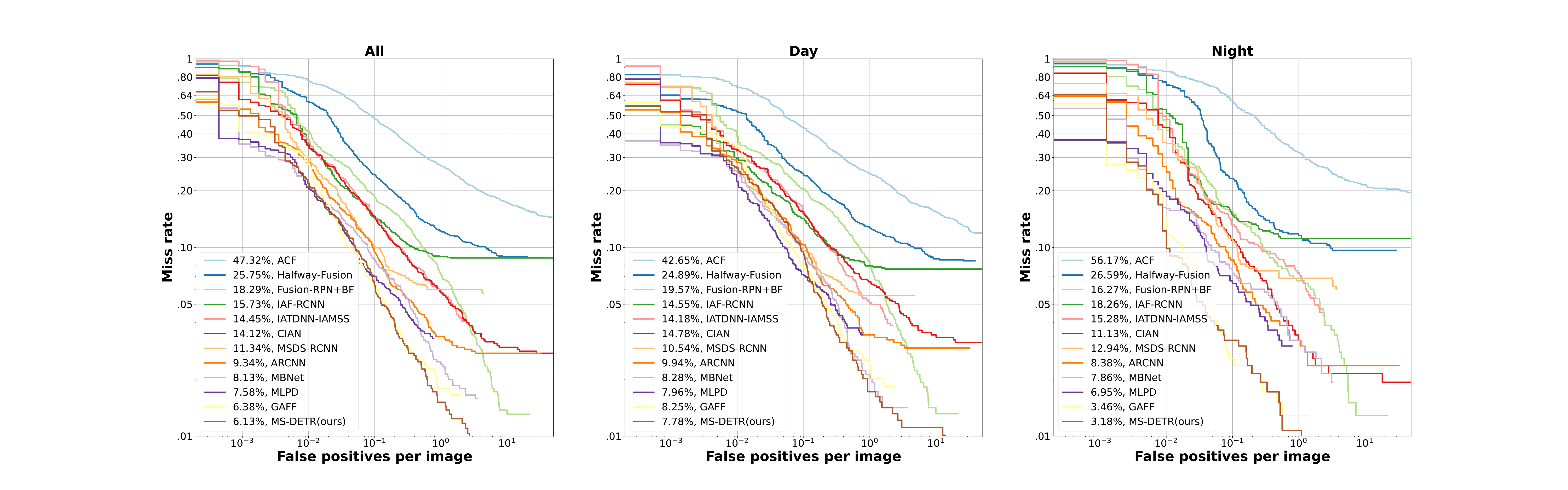}
  \caption{The FPPI-MR curces comparisons with the state-of-the-art methods on the KAIST dataset under ``\textit{Reasonable}" setting.}
  \label{fig:FPPI_Reasonable}
\end{figure*}

\begin{figure*}[t]
  \centering
  \includegraphics[width=0.9\textwidth]{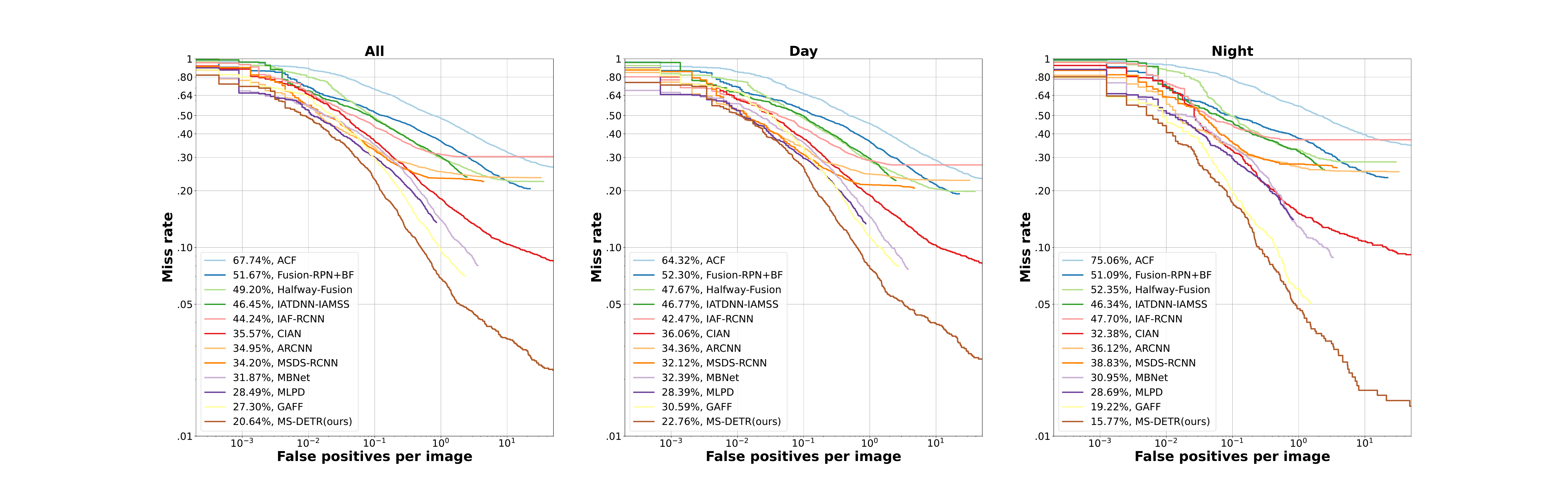}
  \caption{The FPPI-MR curces comparisons with the state-of-the-art methods on the KAIST dataset under ``\textit{All}" setting.}
  \label{fig:FPPI_All}
\end{figure*}

\subsection{Implementation Details}
Since DETR-like detectors are generally data-hungry~\cite{wang2022towards}, we pretrain our model on COCO dataset. 
In addition to random resize, flip and affine, we also adopt strong data augmentation, including Mixup~\cite{zhang2017mixup}, Mosaic~\cite{bochkovskiy2020yolov4} and RandAugment~\cite{cubuk2020randaugment}. 

The proposed MS-DETR is implemented with Pytorch~\cite{paszke2019pytorch}.
The backbones for both visible and thermal branches are ResNet50~\cite{he2016deep}, and the number of feature scales are $L=4$. Both Transformer encoder and decoder have six layers. MS-DETR has 76.38M parameters in total, and the amount of floating-point operations is 258.84 GFLOPs for an input image with the size of 640$\times$512. 

In our multi-modal decoder, the number of attention heads, sampling points, and object queries are set to be $H=8$, $K =4$ and $N=300$.  
The model is trained using the Adam optimizer~\cite{loshchilov2017decoupled}. We train it twenty epochs for KAIST and LLVIP datasets, and ten epoches for CVC-14 dataset. 
The learning rate is initialized as 0.0001 and is decayed at the half of the training process by a factor of ten. When we train our model on four RTX-3090 GPUs with a batch size of 2, the training lasts approximately 7 hours. The inference time is approximately 0.11 seconds per image.


\subsection{Comparisons with State-of-the-art}

\textbf{KAIST.} 
We compare our model on KAIST dataset with several state-of-the-art methods, including ACF~\cite{KAIST}, Halfway Fusion~\cite{liu2016multispectral}, Fusion RPN+BF~\cite{konig2017fully}, IAF R-CNN~\cite{li2019illumination}, IATDNN+IASS~\cite{guan2019fusion}, CIAN~\cite{zhang2019cross}, MSDS-RCNN~\cite{limultispectral}, AR-CNN~\cite{zhang2019weakly},  MBNet~\cite{zhou2020improving}, CMPD~\cite{li2022confidence}, 
Faster RCNN+MFPT~\cite{zhu2023multi}, MLPD~\cite{kim2021mlpd}, GAFF~\cite{zhang2021guided}, ProbEn$_3$~\cite{chen2022multimodal}, MCHE-CF~\cite{mch-li2023} and AANet-Faster RCNN~\cite{aanet-2023}. Table~\ref{tab:KAIST} provides the performance comparisons. We can observe our proposed model achieves the best performance under both ``\textit{Reasonable}" and ``\textit{All}" settings, especially under ``\textit{All}" setting, where our MS-DETR outperformed GAFF~\cite{zhang2021guided} by 6.66\%. Note that the ``\textit{All}" setting is more difficult since it contains small and heavy occluded pedestrians. This performance contrast illustrates that our method can well detect small pedestrians. To validate this, we further evaluate our method on the six subsets in terms of pedestrian distance and occlusion level, whose results are shown in Table \ref{tab:KAISTsubset}. 
In this table, we can see our method largely reduce the miss rate in almost all cases, except for heavy occluded pedestrians. Note that MS-DETR performs the second best in Heavy subset, with a slightly higher $MR^{-2}$ than GAFF~\cite{zhang2021guided}. Actually, almost all comparison methods do not perform well on the Heavy subset. Because more than 50$\%$ of the pedestrian area are occluded, only weaker features can be learned. In GAFF~\cite{zhang2021guided}, the pedestrian mask is supervised by a groundtruth mask generated from the bounding box annotation, and the visible and thermal features are enhanced with the guidance of segmentation masks. We believe this pixel-level supervision truly enhances the features and helps the model to concentrate more on the pedestrians even in the scenarios with severe occlusions.

Moreover, the FPPI-MR curves on the ``\textit{Reasonable}" and ``\textit{All}" settings are also demonstrated in Fig.~\ref{fig:FPPI_Reasonable} and Fig.~\ref{fig:FPPI_All}. Under the ``Reasonable" setting, our model shows great improvements in the night scenarios,  
where it obtains the lowest miss rate in a wide range. Furthermore, we can observe from Fig.~\ref{fig:FPPI_All} that MS-DETR has obvious advantages in most FPPI ranges compared to all comparison methods both in daytime and nighttime scenarios. To be specific, our model demonstrates larger superiority than other comparisons at a bigger FPPI under both ``\textit{Reasonable}" and ``\textit{All}" settings.

\begin{figure*}[t]
  \centering
  \includegraphics[width=1\textwidth]{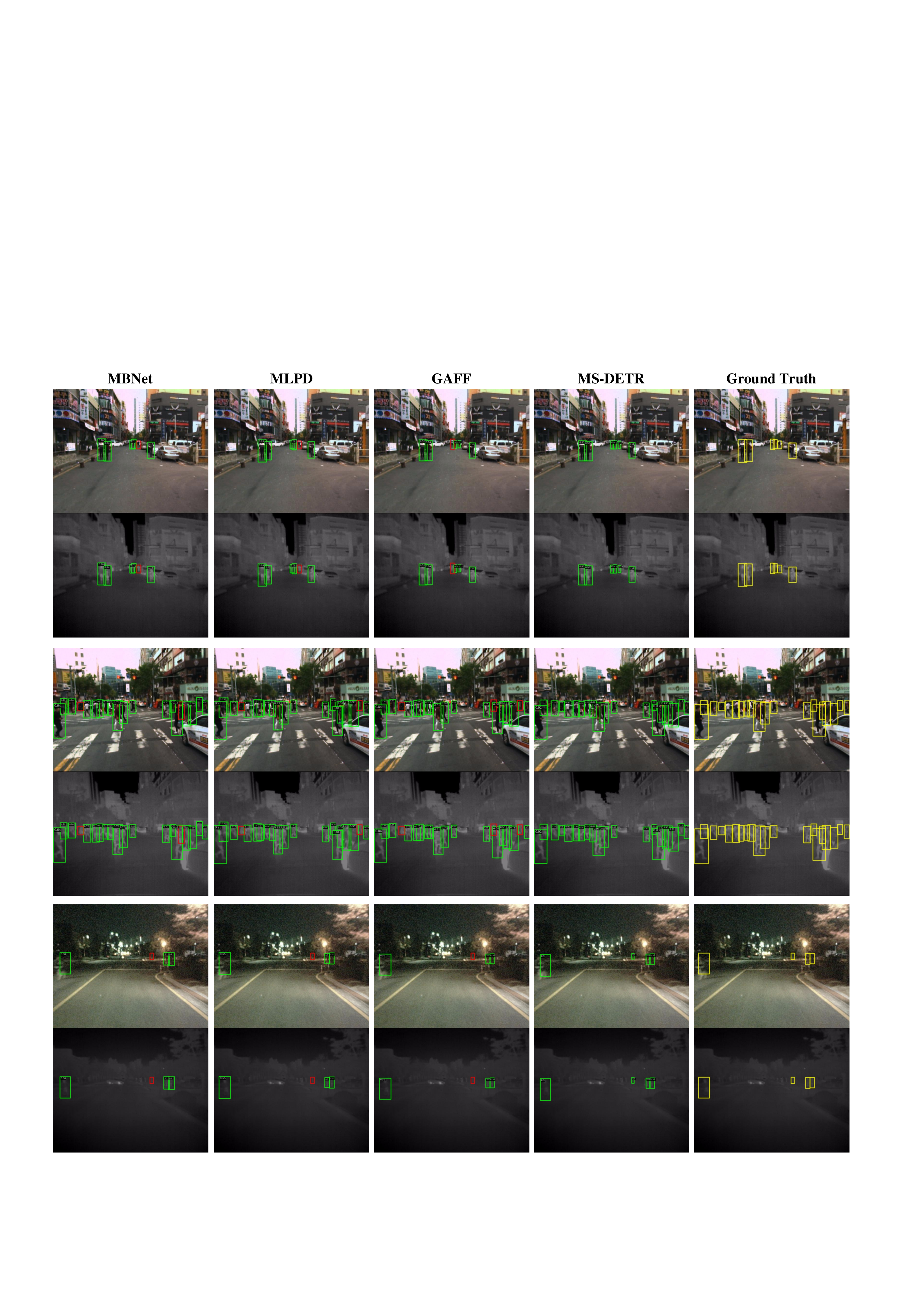}
  \caption{Some detection results of our model and other three state-of-the-art models on KAIST dataset. The groundtruths, predictions and missed predictions of different detectors are marked in yellow, green and red boxes. Our MS-DETR demonstrates superior results with less missed detections.}
  \label{fig:KAISTsample}
\end{figure*}

Finally, we visualize some detection results in Fig.~\ref{fig:KAISTsample}, where the yellow rectangles locate the ground truth pedestrians, the green rectangles represent the predictions of MBNet~\cite{zhou2020improving}, MLPD~\cite{kim2021mlpd}, GAFF~\cite{zhang2021guided} and MS-DETR, and red rectangles are missed detection samples of these detectors. From these figures, we can observe that there are some pedestrians missed by the compared methods. In contrast, our model can well locate these challenging samples, especially these small pedestrians. Furthermore, the positive sample confidence scores output by our model is relatively higher than others. We contribute the robust detection of pedestrians to our loosely coupled fusion strategy, which can eliminate the interference of occlusion by fusing the prominent key elements. 

\begin{figure}[t]
  \includegraphics[scale=0.21]{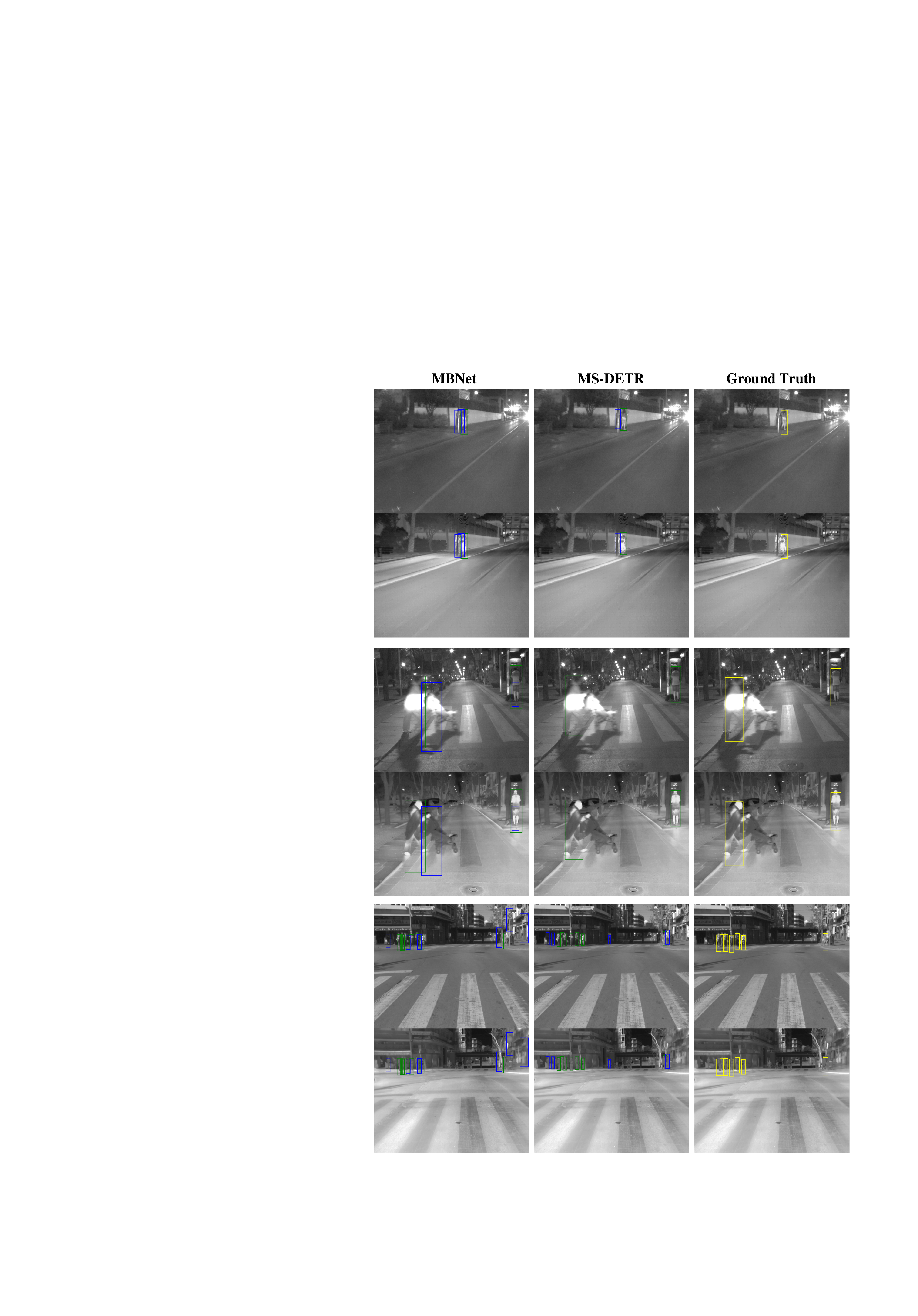}
  \caption{Some detection results on CVC-14 dataset. The groundtruths, correct predictions, and false positives are marked in yellow, green, and blue boxes, respectively. The selected scenes represent typical examples of crowded, close-up, and occluded situations. It is evident that MS-DETR exhibits fewer false positives and detects pedestrians more accurately.}
  \label{fig:cvc_vis}
\end{figure}

\textbf{CVC-14.} Our model is compared with MACF~\cite{park2018unified}, Choi \etal~\cite{choi2016multi}, Halfway Fusion~\cite{liu2016multispectral}, Park \etal~\cite{park2018unified}, AR-CNN~\cite{zhang2019weakly}, MBNet~\cite{zhou2020improving}, CMPD~\cite{li2022confidence}, MLPD~\cite{kim2021mlpd}, MCHE-CF~\cite{mch-li2023} and AANet-Faster RCNN~\cite{aanet-2023} 
on the CVC-14 dataset. Like \cite{zhou2020improving}, we fine-tune the model pretrained on the KAIST dataset using the CVC-14  to further boost the performance. Due to the serious misalignment between visible and thermal modalities in CVC-14 dataset, we also adopt the strategy in \cite{zhang2019weakly}, where the pedestrians in the visible modality as the training target and the pedestrians in thermal modality as a reference. The final results using dual-modalities are listed in Table~\ref{tab:CVC14}.
From the table, we can observe that MS-DETR performs well under ``All" and ``Night" settings.
As for the performance under the ``Day" setting, we conjecture that the suboptimal result of MS-DETR is attributed to the fact that the loosely coupled fusion strategy only considers selected points. This sparse selection process loses some texture information, which is vital for visible images, especially in daytime scenes.
Note that our model does not have any well-designed alignment modules or operations, but it can also obtain lower miss rate compared with other models, especially in night scenario.

Some visualizations are provided in Fig.~\ref{fig:cvc_vis}, where the groundtruthes, true positives, and 
false positive boxes are shown in yellow, green, and blue colors, respectively. We observe that MS-DETR has fewer false positives and, in some scenarios, can detect more pedestrians on the challenging CVC-14 dataset, despite its partial annotations. At the same time, it shows robustness to occlusion. 
In contrast, MBNet has more false positives.
\begin{table}[t!]\footnotesize
    \caption{Performance comparisons on CVC-14 dataset}
    \centering
    \begin{tabular}{c|c|c|c}
        \hline
        \multirow{2}{*}{Methods}              & \multicolumn{3}{c}{$MR^{-2}$(IoU=0.5)} \\ \cline{2-4} 
                                              & All  & Day  & Night                    \\ \hline
        MACF~\cite{park2018unified}           & 60.1 & 61.3 & 48.2                     \\
        Choi \etal~\cite{choi2016multi}       & 47.3 & 49.3 & 43.8                     \\
        Halfway Fusion~\cite{park2018unified} & 37.0 & 38.1 & 34.4                     \\
        Park \etal~\cite{park2018unified}     & 31.4 & 31.8 & 30.8                     \\
        AR-CNN~\cite{zhang2019weakly}         & 22.1 & 24.7 & 18.1                     \\
        MCHE-CF~\cite{mch-li2023}              & 21.3 & 24.0 & 17.5  \\
        MLPD~\cite{kim2021mlpd}               & 21.3 & 24.2 & 18.0                     \\
        MBNet*~\cite{zhou2020improving}        & 21.1 & 24.7 & 13.5                     \\
        AANet-Faster RCNN*~\cite{aanet-2023}    & \underline{17.9} & \textbf{22.6} & \underline{10.9}  \\ \hline
        MS-DETR                                 & 20.0 & \underline{23.0} & 15.1    \\
        MS-DETR*                               & \textbf{16.9} & 24.1 & \textbf{8.8}    \\ \hline
    \end{tabular}
    \begin{tablenotes}
    \item[*] * indicates that the model is clearly announced to be fine-tuned from the KAIST pretrained model.
    \end{tablenotes}
    \label{tab:CVC14}
\end{table}

\textbf{LLVIP.} For LLVIP dataset, we compare our method with DCMNet~\cite{xie2022learning}, CSAA~\cite{cao2023multimodal}, CFT~\cite{qingyun2021cross} and LRAF-Net~\cite{fu2023lraf}. The results are provided in Table~\ref{tab:LLVIP}. 
Note that we also present the experimental results of single-modality models. Since LLVIP is a dataset of night scenes, the detection results on thermal modality are generally higher than that on visible modality. In the multi-modality case, our model achieves the comparable performance as LRAF-Net~\cite{fu2023lraf} and significantly outperforms other comparison methods. It should be noted that even our single-modality baselines, like the thermal-modality baseline, can obtain superior results than multi-modal comparison methods, which demonstrates the potential of DETR-like models.

\begin{table}[t!]\footnotesize
    \caption{Comparisons with the State-of-the-art Methods on LLVIP Dataset}
    \centering
    \begin{tabular}{c|c|c|c|c}
        \hline
        Methods & Data & $AP_{0.5}$ & $AP_{0.75}$ & $AP$    \\ \hline
                                     \multicolumn{5}{c}{single-modality}   \\ \hline
        YOLOv3~\cite{jia2021llvip}   & V    & 85.9 & 37.9 & 43.3  \\
        YOLOv5~\cite{jia2021llvip}   & V    & 90.8 & 51.9 & 50.0  \\ 
        Ours                            & V    & 91.8 & 50.9 & 50.8  \\ \hline
        YOLOv3~\cite{jia2021llvip}   & T    & 89.7 & 53.4 & 52.8  \\
        YOLOv5~\cite{jia2021llvip}   & T    & 94.6 & 72.2 & 61.9  \\ 
        Ours                            & T    & 97.5 & 75.1 & 65.3  \\ \hline
                                     \multicolumn{5}{c}{multi-modality} \\ \hline
        Faster RCNN + DCMNet~\cite{xie2022learning}  & V+T & - & - & 58.4 \\
        Cascade RCNN + DCMNet~\cite{xie2022learning} & V+T & - & - & 61.5 \\
        RetinaNet + DCMNet~\cite{xie2022learning}    & V+T & - & - & 58.9 \\
        SABL + DCMNet~\cite{xie2022learning}         & V+T & - & - & 60.6 \\
        Reppoints + DCMNet~\cite{xie2022learning}    & V+T & - & - & 58.7 \\
        CSAA~\cite{cao2023multimodal}                                 & V+T & 94.3 & 66.6 & 59.2 \\
        CFT\cite{qingyun2021cross}                      & V+T & \underline{97.5}  & \underline{72.9}  & 63.6   \\
        LRAF-Net~\cite{fu2023lraf}              & V+T & \textbf{97.9}  & -     & \textbf{66.3}   \\ \hline
        MS-DETR                                      & V+T & \textbf{97.9}  & \textbf{76.3}  & \underline{66.1}   \\ \hline
    \end{tabular}
    \label{tab:LLVIP}
\end{table}

\subsection{Ablation Studies}
To investigate the effectiveness of our proposed loosely coupled fusion and modality-balanced optimization, we conduct some ablation studies. For simplicity, we only conduct experiments on KAIST dataset and provide the miss rate under both the ``\textit{Reasonable}'' and ``\textit{All}'' settings. 

\textbf{Effectiveness of loosely coupled fusion strategy.}  
We conduct a series of experiments, including the single-modal detection and the multi-modal detection with different fusion strategies, to verify the effectiveness of proposed loosely coupled fusion. 
We can firstly observe from Table~\ref{tab:ablation} that multi-modal detection shows better performance in detecting pedestrians than single-modality by a large margin. Then we investigate the influence of different fusion strategies. 

In~\cite{liu2016multispectral}, the authors divided the fusion methods into early fusion, halfway fusion and late fusion according to the stage at which the feature fusion operation is performed. 
Similarly, our CNN backbone, i.e., ResNet-50, has five stages, named stage-1 to stage-5. Taking the whole Transformer encoder as the last stage, we totally have six stages. Different fusion strategies are defined as follows: 
\begin{itemize}
    \item \textbf{Early fusion} begins fusion  after stage-1 and shares network parameters from stage-2 to stage-6.
    \item \textbf{Halfway fusion\#1} performs fusion from stage-3 to stage-5 and shares networks parameters of stage-6. \item \textbf{Halfway fusion\#2} is performed after stage-3 and shares from stage-4 to stage-6. 
    \item \textbf{Encoder fusion} is performed in Transformer encoder, i.e. stage-6, using the deformable self-attention module.
    \item \textbf{Late fusion} only fuses features after stage-6.
\end{itemize}
 As for the fusion process, we first concatenate corresponding features, and then use a $3 \times 3$ convolutional layer, followed by batch-normalization and ReLU activation, to obtain the fused features. The results are provided in Table~\ref{tab:ablation}. When comparing different fusion strategies, we find that late fusion has better performance, and our proposed loosely coupled fusion brings the drop of $MR^{-2}$ from 0.77 to 3.49, which validates its effectiveness in DETR-like detection models.

The ablation experiment on CVC-14 dataset is also shown in Table~\ref{tab:ablation}. 
From the ablations, we can observe that our proposed loosely coupled fusion and the modality-balanced optimization strategies show great potential on 
CVC-14 dataset.
We believe this may be due to the large number of misaligned samples in the CVC-14 dataset. While the dataset provides annotations for both modalities, we used only the visible modality annotations for training. Consequently, the trained model naturally leans towards the visible modality, making the inclusion of loosely coupled fusion and modality-balanced optimization particularly critical.

\begin{table*}[t]\footnotesize
    \caption{Ablation Studies. (
    \textit{MBO: Modality-Balanced Optimization. V: Visible. T: Thermal})}
    \centering
    \renewcommand\arraystretch{1.2}
    \tabcolsep=0.01cm
    \setlength{\tabcolsep}{3mm}{
        \begin{tabular}{c | c | c c c c c | c | c | c }
        \hline
        {Dataset} &
        {Modality} & \multicolumn{6}{c|}{Fusion Strategy} & \quad & {Metrics} \\
        \hline
            \multirow{10}{*}{KAIST}    & \quad          & Early & Halfway \#1 & Halfway \#2 & Encoder & Late & Loosely Coupled   & MBO   & R-All          \\  \cline{2-10}
            \quad & V      & \quad        & \quad       & \quad      & \quad       & \quad       & \quad      & \quad & 18.18       \\ \cline{2-10}
            \quad   & T & \quad        & \quad       & \quad      & \quad       & \quad       & \quad      & \quad & 15.61      \\ \cline{2-10}
            \quad & \multirow{7}{*}{V+T} & \checkmark   & \quad       & \quad      & \quad       & \quad       & \quad      & \quad & 10.75      \\
            \quad & \quad & \quad        & \checkmark  & \quad      & \quad       & \quad       & \quad      & \quad & 8.60        \\
            \quad & \quad & \quad        & \quad       & \checkmark & \quad       & \quad       & \quad      & \quad & 7.96       \\
            \quad & \quad & \quad        & \quad       & \quad      & \checkmark  & \quad       & \quad      & \quad & 8.70       \\
            \quad & \quad & \quad        & \quad       & \quad      & \quad       & \checkmark  & \quad      & \quad & 7.56       \\
            \quad & \quad & \quad        & \quad       & \quad      & \quad       & \quad       & \checkmark & \quad & 6.79       \\ 
            \quad & \quad & \quad        & \quad       & \quad      & \quad       & \quad       & \checkmark & \checkmark & \textbf{6.13}  \\ \hline
            \multirow{6}{*}{CVC-14} & \multirow{6}{*}{V+T} & \checkmark   & \quad       & \quad      & \quad       & \quad       & \quad      & \quad & 24.09      \\
            \quad & \quad & \quad        & \checkmark  & \quad      & \quad       & \quad       & \quad      & \quad & 18.90        \\
            \quad & \quad & \quad        & \quad       & \checkmark & \quad       & \quad       & \quad      & \quad & 21.79       \\
            \quad & \quad & \quad        & \quad       & \quad      & \checkmark  & \quad       & \quad      & \quad & 24.31      \\
            \quad & \quad & \quad        & \quad       & \quad      & \quad       & \checkmark  & \quad      & \quad & 23.27      \\
            \quad & \quad & \quad        & \quad       & \quad      & \quad       & \quad       &\checkmark & \checkmark & \textbf{16.90} \\
            
        \hline
        \end{tabular}}
        \label{tab:ablation}
\end{table*}


\begin{table}[t]\footnotesize
    \caption{Performance comparisons within three detection branch on KAIST dataset}
    \centering
    \setlength{\tabcolsep}{2mm}{
        \begin{tabular}{c | c c c | c c c }
        \hline
            Detection Branch & R-All & R-Day & R-Night & All   & Day   & Night  \\ \hline
            Visible          & 6.28  & 7.98  & 3.37    & 20.92 & 23.19 & 15.77    \\
            Thermal          & 6.16  & 7.89  & 3.22    & 20.93 & 23.37 & 15.94    \\
            Fusion           & 6.13  & 7.78  & 3.18    & 20.64 & 22.76 & 15.77        \\ \hline
        \end{tabular}}
    \label{tab:balance}
\end{table}

\begin{table}[t]\footnotesize
    \caption{Effect of the number of encoder layers on KAIST dataset}
    \centering
    \setlength{\tabcolsep}{2mm}{
        \begin{tabular}{c | c c c | c c c }
        \hline
        \multirow{2}{*}{Number of Layers} & \multicolumn{3}{c|}{\textit{Reasonable}}  & \multicolumn{3}{c}{\textit{All}} \\ \cline{2-7}
             & R-All & R-Day & R-Night & All & Day & Night \\   
            \cline{1-7}
            6 & 6.13 & 7.78 & 3.18 & 20.64 & 22.76 & 15.77 \\
            5 & 6.50 & 8.39 & 3.19 & 22.36 & 24.58 & 16.48 \\
            4 & 7.08 & 9.05 & 3.56 & 23.10 & 25.26 & 18.02 \\
            3 & 6.77 & 8.14 & 4.15 & 22.18 & 24.26 & 17.08 \\
            2 & 6.97 & 8.50 & 4.15 & 22.69 & 24.57 & 18.55 \\
            1 & 7.77 & 9.74 & 4.01 & 22.31 & 24.54 & 16.90  \\ 
        \hline
        \end{tabular}}
  \label{tab:encoder_layers}
\end{table}

\begin{table}[t]\footnotesize
    \caption{Effect of the number of decoder layers on KAIST dataset}
    \centering
    \setlength{\tabcolsep}{2mm}{
        \begin{tabular}{  c| c  c  c | c  c  c}
        \hline
        \multirow{2}{*}{Number of Layers} & \multicolumn{3}{c|}{\textit{Reasonable}}  & \multicolumn{3}{c}{\textit{All}} \\ \cline{2-7}
             & R-All & R-Day & R-Night & All & Day & Night \\   \cline{1-7}
            6 & 6.13 & 7.78 & 3.18 & 20.64 & 22.76 & 15.77 \\
            5 & 6.89 & 8.09 & 4.31 & 22.83 & 24.44 & 19.23 \\
            4 & 7.50 & 9.36 & 4.17 & 22.85 & 25.20 & 17.80 \\
            3 & 8.25 & 10.27 & 4.56 & 23.24 & 24.99 & 19.47 \\
            2 & 10.44 & 12.07 & 7.65 & 26.54 & 28.97 & 20.75 \\
            1 & 22.02 & 24.92 & 15.21 & 37.37 & 40.65 & 29.56 \\ \hline
        \end{tabular}}
  \label{tab:decoder_layers}
\end{table}

\textbf{Effectiveness of modality-balanced optimization strategy.}
To better measure and balance the contribution of visible and thermal modalities, we design an instance-aware Modality-Balanced Optimization (MBO) strategy, where we select the branch that has minimum cost in current iteration as reference branch to re-arrange the permutation of prediction slots, rather than directly take fusion branch as reference. Further, we employ a dynamic loss to fuse different branches using an instance-level weight. It can be observed from Table~\ref{tab:ablation} that the model trained on such a strategy can bring $0.66\%$ drops of $MR^{-2}$ without introducing any additional modules. 
Therefore, we derive that modality-balanced optimization strategy boosts the detection performance by  unifying the permutation of prediction slots and dynamically adjusting the weights of three branches in instance level. 

To investigate its modality balance ability, we present the detection results of all three branches (Fusion, Visible and Thermal) in Table~\ref{tab:balance}. We can find that three branches have balanced detection results under both ``\textit{All}" and ``\textit{Reasonable}" settings, which further proves the effectiveness of MBO strategy.

\begin{table}[t]\footnotesize
    \caption{Effect of the number of queries on KAIST dataset}
    \centering
    \setlength{\tabcolsep}{2mm}{
        \begin{tabular}{  c| c  c  c | c  c  c}
        \hline
        \multirow{2}{*}{Number of Queries} & \multicolumn{3}{c|}{\textit{Reasonable}}  & \multicolumn{3}{c}{\textit{All}} \\ \cline{2-7}
             & R-All & R-Day & R-Night & All & Day & Night \\   \cline{1-7}
            300 & 6.13 & 7.78 & 3.18 & 20.64 & 22.76 & 15.77 \\
            200 & 6.79 & 8.36 & 3.73 & 21.27 & 23.01 & 17.50 \\
            100 & 6.65 & 7.90 & 4.27 & 21.63 & 24.14 & 16.36 \\
            50 & 8.09 & 10.08 & 4.34 & 21.88 & 24.53 & 16.15 \\ \hline
        \end{tabular}}
  \label{tab:queries}
\end{table}

\begin{figure*}[t]
  \centering
  \includegraphics[width=0.95\textwidth]{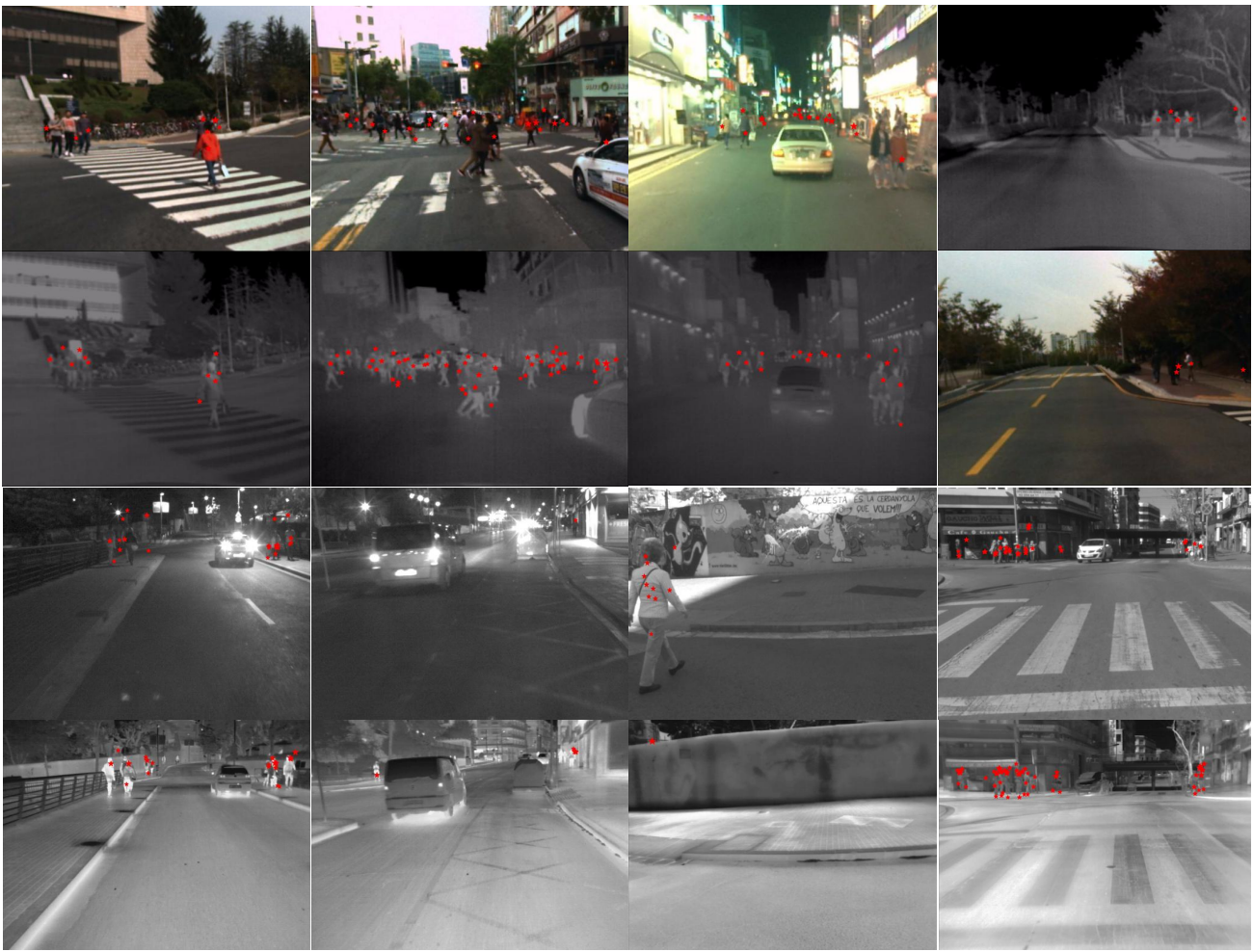}
  \caption{Visualization of loosely coupled fusion strategy, where the first two rows are from the KAIST dataset, and the last two rows are from the CVC-14 dataset. Points are selected and fused with their corresponding weights (whose weights are denoted through the size of stars ``\textcolor{red}{$\bigstar$}"), which 
  demonstrates that the model focuses on the pedestrians within the sample pair. One can zoom in for more details.
  }
  \label{fig:red_star}
\end{figure*}

\textbf{Influence of hyperparameters in MS-DETR.}
To further explore the influence of hyperparameters, \ie, 
the number of encoder layers, decoder layers, and queries in MS-DETR, we conduct a series of experiments. Table~\ref{tab:encoder_layers} provides an ablation study on the number of encoder layers. It shows that as the number of encoder layers increases, the miss rate $MR^{-2}$ also somewhat diminishes. 
In contrast, the number of decoder layers has a more significant effect on model performance, as shown in Table~\ref{tab:decoder_layers}. 
It's evident that as the number of decoder layers decreases, the model's performance significantly diminishes. For instance, with just one decoder layer, the index surges by 15.89\%. This underscores the crucial importance of the decoder design.
Additionally, MS-DETR's performance is expected to improve with an increased number of queries. Table IX presents an ablation study on the number of queries, demonstrating that the model's performance moderately improves as the number of queries increases. Balancing precision and efficiency, we choose to retain the default number of queries in DETR at 300.

\subsection{Visualizations of Loosely Coupled Fusion}
To further demonstrate the effectiveness of the loosely coupled fusion strategy, we visualize the selected points from MS-DETR on the test set. 
We use red stars to represent the selected points, with their size indicating the magnitude of the weights (One can zoom in for a closer look). Note that to ensure overall visibility, we limit the stars from becoming too large. 
The first two rows of Fig.~\ref{fig:red_star} show the selected points from the KAIST dataset, from which we can observe that MS-DETR performs well even in scenarios with crowed pedestrians 
at relatively long distances. The fusion of these accurately located points makes the model concentrate on the most discriminative features of two modalities.

The last two rows of Fig.~\ref{fig:red_star} are from the test set on CVC-14 dataset. 
When severe misalignment exists, especially in the middle two columns, MS-DETR consistently performs well, accurately locating the selected points on the pedestrians.
In scenes where pedestrians are present in one modality, MS-DETR can resist interference from the other modality and select points solely from the current modality.
All the above visualizations demonstrate the effectiveness of the loosely coupled fusion strategy in MS-DETR.

\begin{figure*}[t]
  \centering
  \includegraphics[width=0.95\textwidth]{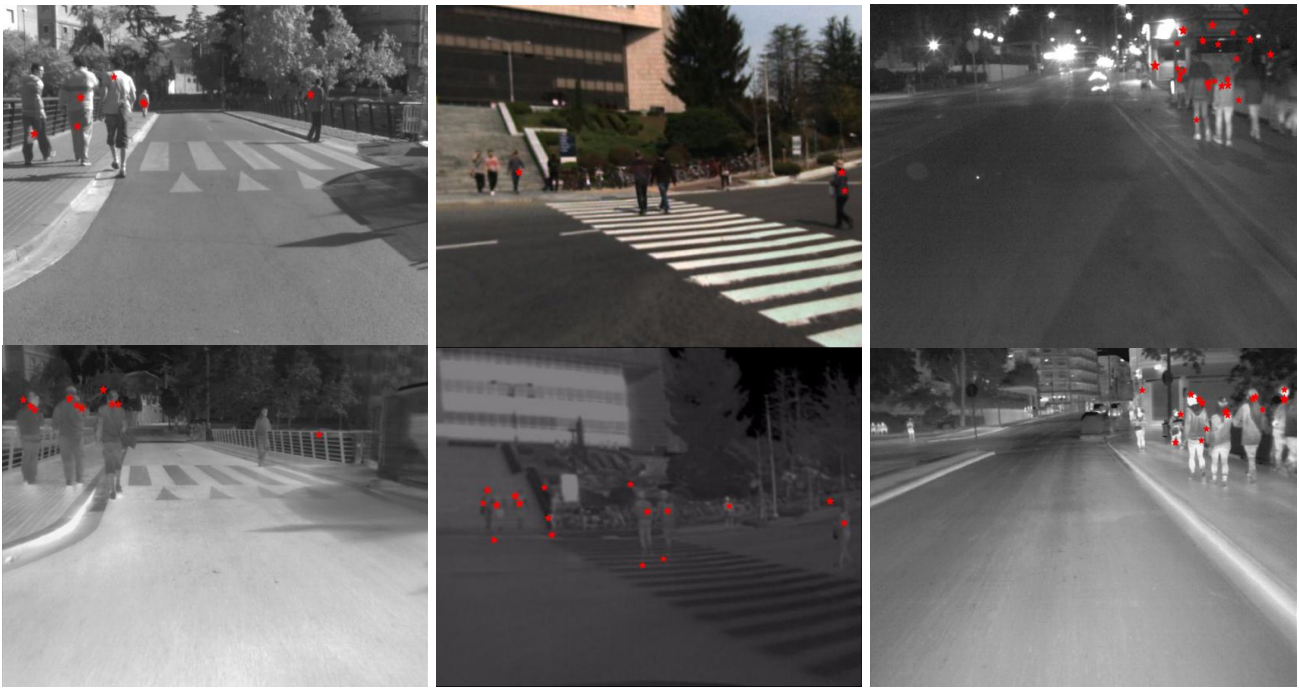}
  \caption{ Visualizations of some typical failure cases on KAIST and CVC-14 datasets, where the first column shows that the selected points may not fall on the same pedestrian due to the large misalignment, the second column demonstrates the number of selected points is imbalanced for two modality images, and the third colunm indicates that some points likely fall on the background even if the pedestrian features are obvious in the visible modality.} 
  \label{fig:fail}
\end{figure*}

\subsection{Further Discussion}
Even though our proposed MS-DETR demonstrates superior detection capability, it still has some limitations. Fig.~\ref{fig:fail} shows some typical failure cases on KAIST and CVC-14 datasets. 
It can be observed from the first column of Fig.~\ref{fig:fail} that the selected points may not fall on the same pedestrian due to the existence of severe misalignment. Since our model sparsely selects prominent points, these biased points directly lead to false positives. Although the loosely coupled fusion has the ability of resisting slight misalignment, the sparse selection mechanism is not enough towards severe misaligned image pairs. To deal with this problem, a registration module that can be combined with our model is essential. 
As for the second column of Fig.~\ref{fig:fail}, MS-DETR selects more points in thermal modality than visible modality even though there are obvious features of pedestrian in the visible modality. 
MS-DETR is designed to adaptively select points from two modality features, as the two modality images contribute differently in daytime or nighttime scenes. However, this flexibility also results in unbalanced feature fusion between the two modalities.
Constraining the number of sampled points may be a feasible solution, but it will sacrifice the flexibility of the model.     
From the third column of Fig.~\ref{fig:fail}, 
MS-DETR sometimes confuses background elements in the visible modality, leading to the false selection of points on the background. The fusion of these false points with selected points from the thermal modality may also trigger false alarms.
From these failure cases, we can conclude that ensuring the accuracy of selected points in the case of severe misalignment, balancing the number of selected points across different modalities, and guaranteeing the adaptability of point selection are crucial areas for further exploration. 

\section{Conclusion}
\label{sec:conclusion}
This paper presents an end-to-end multispectral pedestrian detector (MS-DETR).
A multi-modal Transformer decoder is devised to fuse the visible and thermal features in a loosely coupled fusion way by sparsely sampling some keypoints from the two modalities independently and fusing them with adaptively learned attention weights.
Therefore, the loosely coupled fusion strategy can naturally avoid the misalignment issue.
In addition, by preserving the visible and thermal branches in the decoder, three sets of predicted slots can be obtained and  a novel instance-aware modality-balanced optimization strategy is introduced to measure and adjust the contribution of each instance in visible and thermal images. By a well-designed instance-wise dynamic loss, MS-DETR outperforms existing methods especially on the hard cases, like small or occluded pedestrians.

\textbf{Limitations.} 
While our MS-DETR can resist slight misalignment, it struggles in cases of severe misalignment. In such scenarios, the selected points may be sparse and poorly located on pedestrians, significantly impacting the performance of the loosely coupled fusion strategy. Therefore, considering an early or mid-level fusion of features alongside the loosely coupled strategy may offer a feasible solution. However, dynamically combining multiple fusion strategies remains a challenging and valuable direction for further exploration.
{\small
\bibliographystyle{IEEEtran}
\bibliography{bib}}

\begin{thebibliography}{10}
\providecommand{\url}[1]{#1}
\csname url@samestyle\endcsname
\providecommand{\newblock}{\relax}
\providecommand{\bibinfo}[2]{#2}
\providecommand{\BIBentrySTDinterwordspacing}{\spaceskip=0pt\relax}
\providecommand{\BIBentryALTinterwordstretchfactor}{4}
\providecommand{\BIBentryALTinterwordspacing}{\spaceskip=\fontdimen2\font plus
\BIBentryALTinterwordstretchfactor\fontdimen3\font minus \fontdimen4\font\relax}
\providecommand{\BIBforeignlanguage}[2]{{%
\expandafter\ifx\csname l@#1\endcsname\relax
\typeout{** WARNING: IEEEtran.bst: No hyphenation pattern has been}%
\typeout{** loaded for the language `#1'. Using the pattern for}%
\typeout{** the default language instead.}%
\else
\language=\csname l@#1\endcsname
\fi
#2}}
\providecommand{\BIBdecl}{\relax}
\BIBdecl

\bibitem{enzweiler2008monocular}
M.~Enzweiler and D.~M. Gavrila, ``Monocular pedestrian detection: Survey and experiments,'' \emph{IEEE transactions on pattern analysis and machine intelligence}, vol.~31, no.~12, pp. 2179--2195, 2008.

\bibitem{wang2013scene}
X.~Wang, M.~Wang, and W.~Li, ``Scene-specific pedestrian detection for static video surveillance,'' \emph{IEEE transactions on pattern analysis and machine intelligence}, vol.~36, no.~2, pp. 361--374, 2013.

\bibitem{zhang2016faster}
L.~Zhang, L.~Lin, X.~Liang, and K.~He, ``Is faster r-cnn doing well for pedestrian detection?'' in \emph{European conference on computer vision}.\hskip 1em plus 0.5em minus 0.4em\relax Springer, 2016, pp. 443--457.

\bibitem{mao2017can}
J.~Mao, T.~Xiao, Y.~Jiang, and Z.~Cao, ``What can help pedestrian detection?'' in \emph{Proceedings of the IEEE Conference on Computer Vision and Pattern Recognition}, 2017, pp. 3127--3136.

\bibitem{liu2019high}
W.~Liu, S.~Liao, W.~Ren, W.~Hu, and Y.~Yu, ``High-level semantic feature detection: A new perspective for pedestrian detection,'' in \emph{Proceedings of the IEEE/CVF conference on computer vision and pattern recognition}, 2019, pp. 5187--5196.

\bibitem{abdelmutalab2022pedestrian}
A.~Abdelmutalab and C.~Wang, ``Pedestrian detection using mb-csp model and boosted identity aware non-maximum suppression,'' \emph{IEEE Transactions on Intelligent Transportation Systems}, vol.~23, no.~12, pp. 24\,454--24\,463, 2022.

\bibitem{islam2022pedestrian}
M.~M. Islam, A.~Karimoddini \emph{et~al.}, ``Pedestrian detection for autonomous cars: inference fusion of deep neural networks,'' \emph{IEEE Transactions on Intelligent Transportation Systems}, vol.~23, no.~12, pp. 23\,358--23\,368, 2022.

\bibitem{hsu2023pedestrian}
W.-Y. Hsu and P.-Y. Yang, ``Pedestrian detection using multi-scale structure-enhanced super-resolution,'' \emph{IEEE Transactions on Intelligent Transportation Systems}, 2023.

\bibitem{segments}
C.~Li, W.~Xia, Y.~Yan, B.~Luo, and J.~Tang, ``Segmenting objects in day and night: Edge-conditioned cnn for thermal image semantic segmentation,'' \emph{IEEE Transactions on Neural Networks and Learning Systems}, vol.~32, no.~7, pp. 3069--3082, 2021.

\bibitem{KAIST}
S.~Hwang, J.~Park, N.~Kim, Y.~Choi, and I.~So~Kweon, ``Multispectral pedestrian detection: Benchmark dataset and baseline,'' in \emph{Proceedings of the IEEE conference on computer vision and pattern recognition}, 2015, pp. 1037--1045.

\bibitem{choi2016multi}
H.~Choi, S.~Kim, K.~Park, and K.~Sohn, ``Multi-spectral pedestrian detection based on accumulated object proposal with fully convolutional networks,'' in \emph{2016 23rd International Conference on Pattern Recognition (ICPR)}.\hskip 1em plus 0.5em minus 0.4em\relax IEEE, 2016, pp. 621--626.

\bibitem{konig2017fully}
D.~Konig, M.~Adam, C.~Jarvers, G.~Layher, H.~Neumann, and M.~Teutsch, ``Fully convolutional region proposal networks for multispectral person detection,'' in \emph{Proceedings of the IEEE Conference on Computer Vision and Pattern Recognition Workshops}, 2017, pp. 49--56.

\bibitem{liu2016multispectral}
J.~Liu, S.~Zhang, S.~Wang, and D.~N. Metaxas, ``Multispectral deep neural networks for pedestrian detection,'' in \emph{27th British Machine Vision Conference, BMVC 2016}, 2016.

\bibitem{park2018unified}
K.~Park, S.~Kim, and K.~Sohn, ``Unified multi-spectral pedestrian detection based on probabilistic fusion networks,'' \emph{Pattern Recognition}, vol.~80, pp. 143--155, 2018.

\bibitem{li2019illumination}
C.~Li, D.~Song, R.~Tong, and M.~Tang, ``Illumination-aware faster r-cnn for robust multispectral pedestrian detection,'' \emph{Pattern Recognition}, vol.~85, pp. 161--171, 2019.

\bibitem{guan2019fusion}
D.~Guan, Y.~Cao, J.~Yang, Y.~Cao, and M.~Y. Yang, ``Fusion of multispectral data through illumination-aware deep neural networks for pedestrian detection,'' \emph{Information Fusion}, vol.~50, pp. 148--157, 2019.

\bibitem{zhang2021guided}
H.~Zhang, E.~Fromont, S.~Lef{\`e}vre, and B.~Avignon, ``Guided attentive feature fusion for multispectral pedestrian detection,'' in \emph{Proceedings of the IEEE/CVF winter conference on applications of computer vision}, 2021, pp. 72--80.

\bibitem{dasgupta2022spatio}
K.~Dasgupta, A.~Das, S.~Das, U.~Bhattacharya, and S.~Yogamani, ``Spatio-contextual deep network-based multimodal pedestrian detection for autonomous driving,'' \emph{IEEE transactions on intelligent transportation systems}, vol.~23, no.~9, pp. 15\,940--15\,950, 2022.

\bibitem{zhu2023multi}
Y.~Zhu, X.~Sun, M.~Wang, and H.~Huang, ``Multi-modal feature pyramid transformer for rgb-infrared object detection,'' \emph{IEEE Transactions on Intelligent Transportation Systems}, 2023.

\bibitem{ren2015faster}
S.~Ren, K.~He, R.~Girshick, and J.~Sun, ``Faster r-cnn: Towards real-time object detection with region proposal networks,'' \emph{Advances in neural information processing systems}, vol.~28, 2015.

\bibitem{girshick2014rich}
R.~Girshick, J.~Donahue, T.~Darrell, and J.~Malik, ``Rich feature hierarchies for accurate object detection and semantic segmentation,'' in \emph{Proceedings of the IEEE conference on computer vision and pattern recognition}, 2014, pp. 580--587.

\bibitem{DETR}
N.~Carion, F.~Massa, G.~Synnaeve, N.~Usunier, A.~Kirillov, and S.~Zagoruyko, ``End-to-end object detection with transformers,'' in \emph{European conference on computer vision}.\hskip 1em plus 0.5em minus 0.4em\relax Springer, 2020, pp. 213--229.

\bibitem{peng2022balanced}
X.~Peng, Y.~Wei, A.~Deng, D.~Wang, and D.~Hu, ``Balanced multimodal learning via on-the-fly gradient modulation,'' in \emph{Proceedings of the IEEE/CVF Conference on Computer Vision and Pattern Recognition}, 2022, pp. 8238--8247.

\bibitem{wu2022characterizing}
N.~Wu, S.~Jastrzebski, K.~Cho, and K.~J. Geras, ``Characterizing and overcoming the greedy nature of learning in multi-modal deep neural networks,'' in \emph{International Conference on Machine Learning}.\hskip 1em plus 0.5em minus 0.4em\relax PMLR, 2022, pp. 24\,043--24\,055.

\bibitem{CVC14Dataset}
A.~Gonz{\'a}lez, Z.~Fang, Y.~Socarras, J.~Serrat, D.~V{\'a}zquez, J.~Xu, and A.~M. L{\'o}pez, ``Pedestrian detection at day/night time with visible and fir cameras: A comparison,'' \emph{Sensors}, vol.~16, no.~6, p. 820, 2016.

\bibitem{jia2021llvip}
X.~Jia, C.~Zhu, M.~Li, W.~Tang, and W.~Zhou, ``Llvip: A visible-infrared paired dataset for low-light vision,'' in \emph{Proceedings of the IEEE/CVF international conference on computer vision}, 2021, pp. 3496--3504.

\bibitem{limultispectral}
C.~Li, D.~Song, R.~Tong, and M.~Tang, ``Multispectral pedestrian detection via simultaneous detection and segmentation,'' in \emph{British Machine Vision Conference (BMVC)}.

\bibitem{cao2019box}
Y.~Cao, D.~Guan, Y.~Wu, J.~Yang, Y.~Cao, and M.~Y. Yang, ``Box-level segmentation supervised deep neural networks for accurate and real-time multispectral pedestrian detection,'' \emph{ISPRS journal of photogrammetry and remote sensing}, vol. 150, pp. 70--79, 2019.

\bibitem{li2022confidence}
Q.~Li, C.~Zhang, Q.~Hu, H.~Fu, and P.~Zhu, ``Confidence-aware fusion using dempster-shafer theory for multispectral pedestrian detection,'' \emph{IEEE Transactions on Multimedia}, 2022.

\bibitem{zhang2019weakly}
L.~Zhang, X.~Zhu, X.~Chen, X.~Yang, Z.~Lei, and Z.~Liu, ``Weakly aligned cross-modal learning for multispectral pedestrian detection,'' in \emph{Proceedings of the IEEE/CVF International Conference on Computer Vision}, 2019, pp. 5127--5137.

\bibitem{zhou2020improving}
K.~Zhou, L.~Chen, and X.~Cao, ``Improving multispectral pedestrian detection by addressing modality imbalance problems,'' in \emph{European Conference on Computer Vision}.\hskip 1em plus 0.5em minus 0.4em\relax Springer, 2020, pp. 787--803.

\bibitem{kim2021mlpd}
J.~Kim, H.~Kim, T.~Kim, N.~Kim, and Y.~Choi, ``Mlpd: Multi-label pedestrian detector in multispectral domain,'' \emph{IEEE Robotics and Automation Letters}, vol.~6, no.~4, pp. 7846--7853, 2021.

\bibitem{zhang2019cross}
L.~Zhang, Z.~Liu, S.~Zhang, X.~Yang, H.~Qiao, K.~Huang, and A.~Hussain, ``Cross-modality interactive attention network for multispectral pedestrian detection,'' \emph{Information Fusion}, vol.~50, pp. 20--29, 2019.

\bibitem{zhang2020multispectral}
H.~Zhang, E.~Fromont, S.~Lefevre, and B.~Avignon, ``Multispectral fusion for object detection with cyclic fuse-and-refine blocks,'' in \emph{2020 IEEE International Conference on Image Processing (ICIP)}.\hskip 1em plus 0.5em minus 0.4em\relax IEEE, 2020, pp. 276--280.

\bibitem{xu2017learning}
D.~Xu, W.~Ouyang, E.~Ricci, X.~Wang, and N.~Sebe, ``Learning cross-modal deep representations for robust pedestrian detection,'' in \emph{Proceedings of the IEEE conference on computer vision and pattern recognition}, 2017, pp. 5363--5371.

\bibitem{zhang2021deep}
H.~Zhang, E.~Fromont, S.~Lefevre, and B.~Avignon, ``Deep active learning from multispectral data through cross-modality prediction inconsistency,'' in \emph{2021 IEEE International Conference on Image Processing (ICIP)}.\hskip 1em plus 0.5em minus 0.4em\relax IEEE, 2021, pp. 449--453.

\bibitem{liu2021deep}
T.~Liu, K.-M. Lam, R.~Zhao, and G.~Qiu, ``Deep cross-modal representation learning and distillation for illumination-invariant pedestrian detection,'' \emph{IEEE Transactions on Circuits and Systems for Video Technology}, vol.~32, no.~1, pp. 315--329, 2021.

\bibitem{kim2021uncertainty}
J.~U. Kim, S.~Park, and Y.~M. Ro, ``Uncertainty-guided cross-modal learning for robust multispectral pedestrian detection,'' \emph{IEEE Transactions on Circuits and Systems for Video Technology}, vol.~32, no.~3, pp. 1510--1523, 2021.

\bibitem{chen2023grid}
Z.~Chen, D.~Hong, and H.~Gao, ``Grid network: Feature extraction in anisotropic perspective for hyperspectral image classification,'' \emph{IEEE Geoscience and Remote Sensing Letters}, 2023.

\bibitem{chen2023temporal}
Z.~Chen, Y.~Wang, H.~Gao, Y.~Ding, Q.~Zhong, D.~Hong, and B.~Zhang, ``Temporal difference-guided network for hyperspectral image change detection,'' \emph{International Journal of Remote Sensing}, vol.~44, no.~19, pp. 6033--6059, 2023.

\bibitem{chen2022global}
Z.~Chen, Z.~Lu, H.~Gao, Y.~Zhang, J.~Zhao, D.~Hong, and B.~Zhang, ``Global to local: A hierarchical detection algorithm for hyperspectral image target detection,'' \emph{IEEE Transactions on Geoscience and Remote Sensing}, vol.~60, pp. 1--15, 2022.

\bibitem{chen2023local}
Z.~Chen, G.~Wu, H.~Gao, Y.~Ding, D.~Hong, and B.~Zhang, ``Local aggregation and global attention network for hyperspectral image classification with spectral-induced aligned superpixel segmentation,'' \emph{Expert Systems with Applications}, vol. 232, p. 120828, 2023.

\bibitem{beal2020toward}
J.~Beal, E.~Kim, E.~Tzeng, D.~H. Park, A.~Zhai, and D.~Kislyuk, ``Toward transformer-based object detection,'' \emph{arXiv preprint arXiv:2012.09958}, 2020.

\bibitem{dosovitskiy2020image}
A.~Dosovitskiy, L.~Beyer, A.~Kolesnikov, D.~Weissenborn, X.~Zhai, T.~Unterthiner, M.~Dehghani, M.~Minderer, G.~Heigold, S.~Gelly \emph{et~al.}, ``An image is worth 16x16 words: Transformers for image recognition at scale,'' \emph{arXiv preprint arXiv:2010.11929}, 2020.

\bibitem{fang2021you}
Y.~Fang, B.~Liao, X.~Wang, J.~Fang, J.~Qi, R.~Wu, J.~Niu, and W.~Liu, ``You only look at one sequence: Rethinking transformer in vision through object detection,'' \emph{Advances in Neural Information Processing Systems}, vol.~34, pp. 26\,183--26\,197, 2021.

\bibitem{bulat2022fs}
A.~Bulat, R.~Guerrero, B.~Martinez, and G.~Tzimiropoulos, ``Fs-detr: Few-shot detection transformer with prompting and without re-training,'' \emph{arXiv preprint arXiv:2210.04845}, 2022.

\bibitem{meng2021conditionalDETR}
D.~Meng, X.~Chen, Z.~Fan, G.~Zeng, H.~Li, Y.~Yuan, L.~Sun, and J.~Wang, ``Conditional detr for fast training convergence,'' in \emph{Proceedings of the IEEE/CVF International Conference on Computer Vision}, 2021, pp. 3651--3660.

\bibitem{dai2021upDETR}
Z.~Dai, B.~Cai, Y.~Lin, and J.~Chen, ``Up-detr: Unsupervised pre-training for object detection with transformers,'' in \emph{Proceedings of the IEEE/CVF conference on computer vision and pattern recognition}, 2021, pp. 1601--1610.

\bibitem{dai2021dynamicDETR}
X.~Dai, Y.~Chen, J.~Yang, P.~Zhang, L.~Yuan, and L.~Zhang, ``Dynamic detr: End-to-end object detection with dynamic attention,'' in \emph{Proceedings of the IEEE/CVF International Conference on Computer Vision}, 2021, pp. 2988--2997.

\bibitem{zhu2020deformableDETR}
X.~Zhu, W.~Su, L.~Lu, B.~Li, X.~Wang, and J.~Dai, ``Deformable detr: Deformable transformers for end-to-end object detection,'' in \emph{International Conference on Learning Representations}, 2020.

\bibitem{zhang2022SAMDETR}
G.~Zhang, Z.~Luo, Y.~Yu, K.~Cui, and S.~Lu, ``Accelerating detr convergence via semantic-aligned matching,'' in \emph{Proceedings of the IEEE/CVF Conference on Computer Vision and Pattern Recognition}, 2022, pp. 949--958.

\bibitem{liu2021dabDETR}
S.~Liu, F.~Li, H.~Zhang, X.~Yang, X.~Qi, H.~Su, J.~Zhu, and L.~Zhang, ``Dab-detr: Dynamic anchor boxes are better queries for detr,'' in \emph{International Conference on Learning Representations}, 2021.

\bibitem{kamath2021mdetr}
A.~Kamath, M.~Singh, Y.~LeCun, G.~Synnaeve, I.~Misra, and N.~Carion, ``Mdetr-modulated detection for end-to-end multi-modal understanding,'' in \emph{Proceedings of the IEEE/CVF International Conference on Computer Vision}, 2021, pp. 1780--1790.

\bibitem{maaz2022class}
M.~Maaz, H.~Rasheed, S.~Khan, F.~S. Khan, R.~M. Anwer, and M.-H. Yang, ``Class-agnostic object detection with multi-modal transformer,'' in \emph{The European Conference on Computer Vision. Springer}, 2022.

\bibitem{chu2023mt}
S.-Y. Chu and M.-S. Lee, ``Mt-detr: Robust end-to-end multimodal detection with confidence fusion,'' in \emph{Proceedings of the IEEE/CVF Winter Conference on Applications of Computer Vision}, 2023, pp. 5252--5261.

\bibitem{chen2022multimodal}
Y.-T. Chen, J.~Shi, Z.~Ye, C.~Mertz, D.~Ramanan, and S.~Kong, ``Multimodal object detection via probabilistic ensembling,'' in \emph{European Conference on Computer Vision}.\hskip 1em plus 0.5em minus 0.4em\relax Springer, 2022, pp. 139--158.

\bibitem{aanet-2023}
N.~Chen, J.~Xie, J.~Nie, J.~Cao, Z.~Shao, and Y.~Pang, ``Attentive alignment network for multispectral pedestrian detection,'' in \emph{Proceedings of the 31st ACM international conference on multimedia}, 2023, pp. 3787--3795.

\bibitem{mch-li2023}
R.~Li, J.~Xiang, F.~Sun, Y.~Yuan, L.~Yuan, and S.~Gou, ``Multiscale cross-modal homogeneity enhancement and confidence-aware fusion for multispectral pedestrian detection,'' \emph{IEEE Transactions on Multimedia}, 2023.

\bibitem{dollar2011pedestrian}
P.~Dollar, C.~Wojek, B.~Schiele, and P.~Perona, ``Pedestrian detection: An evaluation of the state of the art,'' \emph{IEEE transactions on pattern analysis and machine intelligence}, vol.~34, no.~4, pp. 743--761, 2011.

\bibitem{wang2022towards}
W.~Wang, J.~Zhang, Y.~Cao, Y.~Shen, and D.~Tao, ``Towards data-efficient detection transformers,'' in \emph{European conference on computer vision}.\hskip 1em plus 0.5em minus 0.4em\relax Springer, 2022, pp. 88--105.

\bibitem{zhang2017mixup}
H.~Zhang, M.~Cisse, Y.~N. Dauphin, and D.~Lopez-Paz, ``mixup: Beyond empirical risk minimization,'' \emph{arXiv preprint arXiv:1710.09412}, 2017.

\bibitem{bochkovskiy2020yolov4}
A.~Bochkovskiy, C.-Y. Wang, and H.-Y.~M. Liao, ``Yolov4: Optimal speed and accuracy of object detection,'' \emph{arXiv preprint arXiv:2004.10934}, 2020.

\bibitem{cubuk2020randaugment}
E.~D. Cubuk, B.~Zoph, J.~Shlens, and Q.~V. Le, ``Randaugment: Practical automated data augmentation with a reduced search space,'' in \emph{Proceedings of the IEEE/CVF conference on computer vision and pattern recognition workshops}, 2020, pp. 702--703.

\bibitem{paszke2019pytorch}
A.~Paszke, S.~Gross, F.~Massa, A.~Lerer, J.~Bradbury, G.~Chanan, T.~Killeen, Z.~Lin, N.~Gimelshein, L.~Antiga \emph{et~al.}, ``Pytorch: An imperative style, high-performance deep learning library,'' \emph{Advances in neural information processing systems}, vol.~32, 2019.

\bibitem{he2016deep}
K.~He, X.~Zhang, S.~Ren, and J.~Sun, ``Deep residual learning for image recognition,'' in \emph{Proceedings of the IEEE conference on computer vision and pattern recognition}, 2016, pp. 770--778.

\bibitem{loshchilov2017decoupled}
I.~Loshchilov and F.~Hutter, ``Decoupled weight decay regularization,'' \emph{arXiv preprint arXiv:1711.05101}, 2017.

\bibitem{xie2022learning}
J.~Xie, R.~M. Anwer, H.~Cholakkal, J.~Nie, J.~Cao, J.~Laaksonen, and F.~S. Khan, ``Learning a dynamic cross-modal network for multispectral pedestrian detection,'' in \emph{Proceedings of the 30th ACM International Conference on Multimedia}, 2022, pp. 4043--4052.

\bibitem{cao2023multimodal}
Y.~Cao, J.~Bin, J.~Hamari, E.~Blasch, and Z.~Liu, ``Multimodal object detection by channel switching and spatial attention,'' in \emph{Proceedings of the IEEE/CVF Conference on Computer Vision and Pattern Recognition}, 2023, pp. 403--411.

\bibitem{qingyun2021cross}
F.~Qingyun, H.~Dapeng, and W.~Zhaokui, ``Cross-modality fusion transformer for multispectral object detection,'' \emph{arXiv preprint arXiv:2111.00273}, 2021.

\bibitem{fu2023lraf}
H.~Fu, S.~Wang, P.~Duan, C.~Xiao, R.~Dian, S.~Li, and Z.~Li, ``Lraf-net: Long-range attention fusion network for visible--infrared object detection,'' \emph{IEEE Transactions on Neural Networks and Learning Systems}, 2023.

\end{thebibliography}

\begin{IEEEbiography}[{\includegraphics[width=1in,height=1.25in,clip,keepaspectratio]{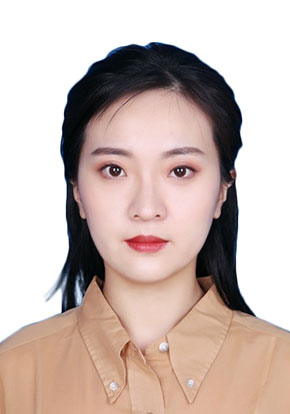}}]{Yinghui Xing}
received a B.S. and Ph.D. degrees from the School of Artificial Intelligence, Xidian University, Xi'an, China, in 2014 and 2020, respectively. She is now an Associate Professor with the School of Computer Science, Northwestern Polytechnical University, Xi'an. Her research interests include remote sensing image processing, image fusion, and infrared target detection.
\end{IEEEbiography}

\vspace{-25pt}

\begin{IEEEbiography}[{\includegraphics[width=1in,height=1.25in,clip,keepaspectratio]{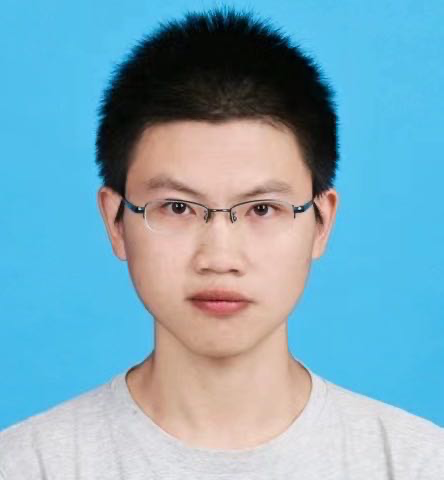}}]{Shuo Yang}
received the B.S. degree from the Taiyuan University of Technology, Shanxi, China, in 2023. He is currently pursuing the M.S. degree with the School of Computer Science, Northwestern Polytechnical University. His research interests include multi-source information fusion and multispectral object detection.
\end{IEEEbiography}

\vspace{-25pt}

\begin{IEEEbiography}[{\includegraphics[width=1in,height=1.25in,clip,keepaspectratio]{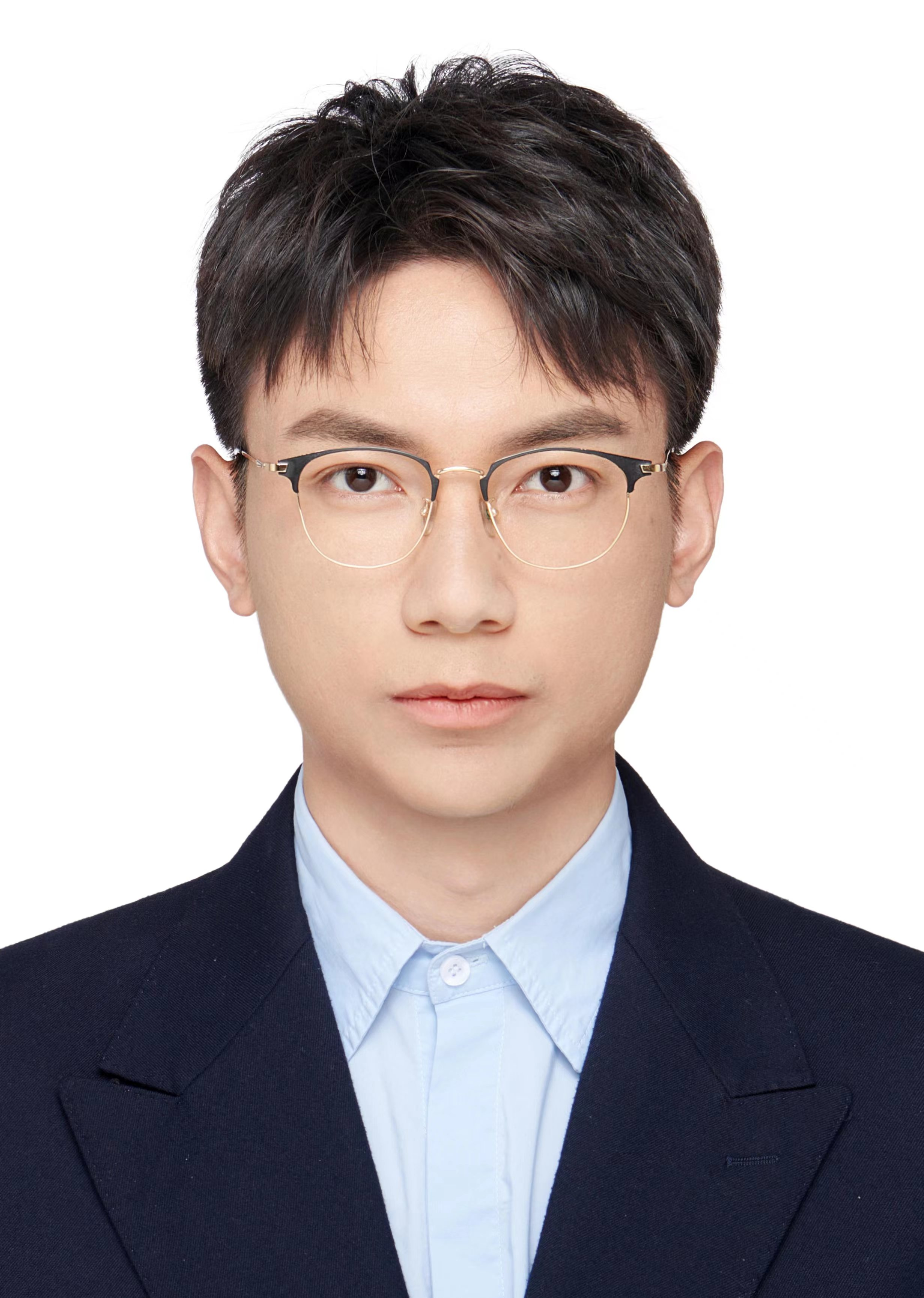}}]{Song Wang}
received the B.S. degree from the Taiyuan University of Technology, Taiyuan, China, in 2018. He is currently pursuing the M.S. degree with the School of Computer Science, Northwestern Polytechnical University. His research interests include multimodal fusion and object detection.
\end{IEEEbiography}

\vspace{-25pt}

\begin{IEEEbiography}[{\includegraphics[width=1in,height=1.25in,clip,keepaspectratio]{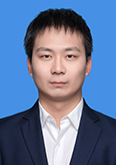}}]{Shizhou Zhang}
received a B.E. and Ph.D. degree from Xi'an Jiaotong University, Xi'an, China, in 2010 and 2017, respectively. Currently, he is with Northwestern Polytechnical University as an Associate Professor (Tenured). His research interests include content-based image analysis, pattern recognition and machine learning, specifically in the areas of deep learning-based vision tasks such as image classification, object detection, reidentification and neural architecture search.

\end{IEEEbiography}

\vspace{-25pt}

\begin{IEEEbiography}[{\includegraphics[width=1in,height=1.25in,clip,keepaspectratio]{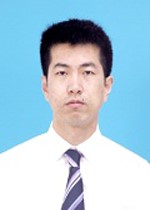}}]{Guoqiang Liang}
received a B.S. in automation and a Ph.D. degrees in pattern recognition and intelligent systems from Xi$'$an Jiaotong University (XJTU), Xi$'$an, China, in 2012 and 2018, respectively. From Mar. to Sep. 2017, he was a visiting Ph.D. Student with the University of South Carolina, Columbia, SC, USA. Currently, he is doing PostDoctoral Research at the School of Computer Science and Engineering, Northwestern Polytechnical University, Xi$'$an, China. His research interests include human pose estimation and human action classification.
\end{IEEEbiography}

\vspace{-25pt}

\begin{IEEEbiography}[{\includegraphics[width=1in,height=1.25in,clip,keepaspectratio]{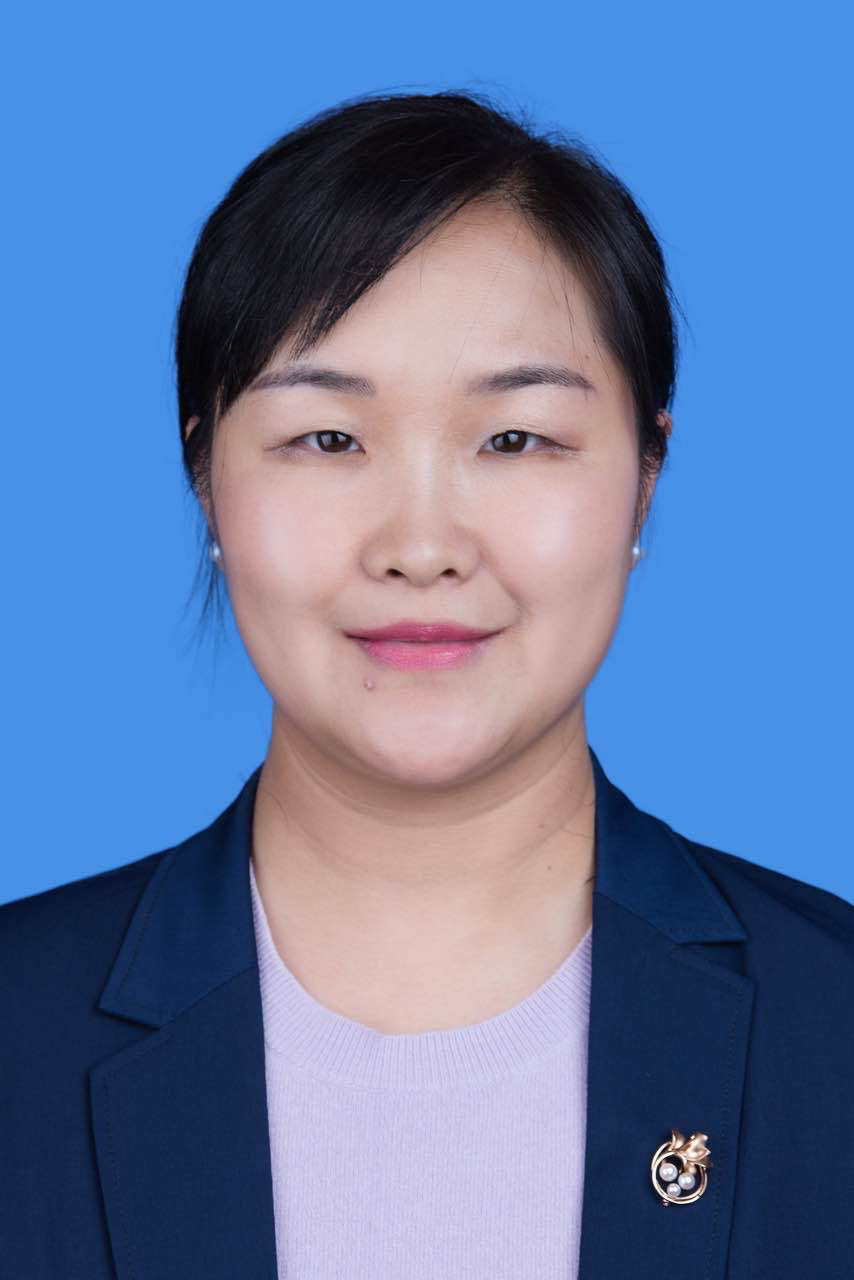}}]{Xiuwei Zhang} received the B.S., M.S. and Ph.D. degrees from the School of Computer Science, Northwestern Polytechnical University, Xi’an, China, in 2004, 2007 and 2011, respectively. She is now an Associate Professor with the School of Computer Science, Northwestern Polytechnical University, Xi’an. 
Her research interests include remote sensing image processing, multi-model image fusion, image registration and intelligent forecasting.
\end{IEEEbiography}

\vspace{-25pt}

\begin{IEEEbiography}[{\includegraphics[width=1in,height=1.25in,clip,keepaspectratio]{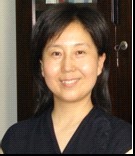}}]{Yanning Zhang}
(SM'10) received her B.S. degree from Dalian University of Science and Engineering in 1988, M.S. and Ph.D. degrees from Northwestern Polytechnical University in 1993 and 1996, respectively. She is presently a Professor at the School of Computer Science, Northwestern Polytechnical University. She is also the organization chair of the Ninth Asian Conference on Computer Vision (ACCV2009). Her research work focuses on signal and image processing, computer vision and pattern recognition. She has published over 200 papers in international journals, conferences and Chinese key journals.
\end{IEEEbiography}

\vfill

\end{document}